# A Survey of Neural Network Robustness Assessment in Image Recognition

Jie Wang

The Key Laboratory on Reliability and Environment Engineering Technology, School of Reliability and Systems Engineering, Beihang University, wang_jie@buaa.edu.cn

Jun Ai

The Key Laboratory on Reliability and Environment Engineering Technology, School of Reliability and Systems Engineering, Beihang University, aijun@buaa.edu.cn

Minyan Lu

The Key Laboratory on Reliability and Environment Engineering Technology, School of Reliability and Systems Engineering, Beihang University, lmy@buaa.edu.cn

Haoran Su

The Key Laboratory on Reliability and Environment Engineering Technology, School of Reliability and Systems Engineering, Beihang University, suhaoran@buaa.edu.cn

Dan Yu

The Key Laboratory on Reliability and Environment Engineering Technology, School of Reliability and Systems Engineering, Beihang University, 18376206@buaa.edu.cn

Yutao Zhang

The Key Laboratory on Reliability and Environment Engineering Technology, School of Reliability and Systems Engineering, Beihang University, zzt7268@gmail.com

Junda Zhu

The Key Laboratory on Reliability and Environment Engineering Technology, School of Reliability and Systems Engineering, Beihang University, 19375379@buaa.edu.cn

Jingyu Liu

The Key Laboratory on Reliability and Environment Engineering Technology, School of Reliability and Systems Engineering, Beihang University, liujingyu1@buaa.edu.cn

In recent years, there has been significant attention given to the robustness assessment of neural networks. Robustness plays a critical role in ensuring reliable operation of artificial intelligence (AI) systems in complex and uncertain environments. Deep learning's robustness problem is particularly significant, highlighted by the discovery of adversarial attacks on image classification models. Researchers have dedicated efforts to evaluate robustness in diverse perturbation conditions for image recognition tasks. Robustness assessment encompasses two main techniques: robustness verification/ certification for deliberate adversarial attacks and robustness testing for random data corruptions. In this survey, we present a detailed examination of both adversarial robustness (AR) and corruption robustness (CR) in neural network assessment. Analyzing current research papers and standards, we provide an extensive overview of robustness assessment in image recognition. Three essential aspects are analyzed: concepts, metrics, and assessment methods. We investigate the perturbation metrics and range representations used to measure the degree of perturbations on images, as well as the robustness metrics specifically for the robustness conditions of classification models. The strengths and limitations of the existing methods are also discussed, and some potential directions for future research are provided.

*Keywords*: robustness assessment, neural network verification, neural network testing, adversarial attacks, data corruptions, image recognition

# 1 INTRODUCTION

Deep learning has achieved remarkable success in computer vision tasks such as image recognition, object detection, autonomous driving, and medical imaging analysis. However, as deep neural networks (DNNs) are increasingly deployed in safety- and security-critical applications, ensuring the quality and reliability of these models has become a pressing concern. Deep learning introduces new failure mechanisms and modes to traditional systems, presenting challenges in evaluating and assuring the quality of intelligent systems. In addition to failures caused by code defects, inadequate prediction and decision-making capabilities due to limited training data, pose significant constraints on their reliable operation in complex and dynamic real-world environments. This deviation between the training and operating environments, known as dataset shift, gives rise to the robustness problem in deep learning. Robustness has thus emerged as a crucial quality characteristic for trustworthy AI. It requires AI systems to exhibit robust behavior throughout their lifecycle, operating normally and posing no unreasonable safety risks under normal use, foreseeable use or misuse, or other unfavorable conditions [1]. In 2022, the ISO 25000 series, with the publication of ISO/IEC DIS 25059:2022 [2], recognizes the importance of robustness as a sub-characteristic of reliability in the quality model for AI systems. The nonlinear and nonconvex behavior of deep neural networks makes their robustness problem serious and difficult to evaluate. The presence of adversarial samples in image classification neural networks highlighted the vulnerability of deep learning models to small input perturbations, which can lead to significant output deviations [3]. In addition to adversarial attacks, neural network models also suffer from robustness problems when operating in natural environments, where there are various kinds of data corruption. A new research topic focuses on robustness assessment of neural networks in the face of real-world data corruption perturbations. This corruption robustness (CR) differs from earlier studies on adversarial robustness (AR) according to the definitions in [4]. The CR specifically refers to unintended changes in data. In the field of computer science, data corruption is "errors in computer data that occur during writing, reading, storage, transmission, or processing, which introduce unintended changes to the original data. " [5]. In the physical environment, typical data corruption perturbations for images include Gaussian blur, rain and snow-induced blur, brightness variations, spatial flipping, and signal transmission distortion.

Two main types of assessment methods are employed for evaluating DNN robustness: robustness verification and robustness testing. Robustness verification aims to determine if a network is robust within a specific perturbation range and has made significant progress through formal and statistical verification techniques. Robustness testing involves constructing test datasets with perturbations to evaluate the correctness of a model's output to these perturbed inputs. Current research on robustness testing primarily focuses on addressing issues such as test input generation, test oracle generation, and test adequacy analysis. Existing methods are primarily proposed for adversarial attacks and AR, aiming to identify the minimum perturbation degree that misleads the model output and serve it as the measurement of robustness. The assessment of CR often adopts a benchmark testing approach, which requires the construction of effective and widely recognized benchmarks for robustness evaluation.

While some surveys [6, 7] provide reviews on the development and current state of AR, there is a need for a comprehensive summary that clarifies the relationship between AR, CR, and other existing robustness concepts, as well as the assessment methods and metrics for each type of robustness. We aim to bridge this gap by conducting a systematic analysis of the latest progress of neural network robustness assessment encompassing both AR and CR within the domain of image recognition. In total, more than 3,000 relevant documents were searched, with the time periods spanning from 1975 to 2023. Over 400 papers related to the topics of robustness verification, testing, and assessment were analyzed, including 30 review papers and 15 standards. The total number of references cited in this paper is 168. In this survey, we provide a comprehensive overview of the research conducted on the quantitative assessment of neural network robustness in terms of concepts, metrics, and assessment methods. Our work makes the following key contributions:

- **Robustness concepts.** We conduct a detailed analysis of the robustness concepts defined for AI systems and neural networks as per existing standards and research papers. We examine the interplay between robustness and other essential AI quality characteristics. The definition of robustness for neural networks aligns with the system-level definition but more detailed. The existing concepts that describe or measure the robustness of neural networks from different perspectives is discussed. By organizing these concepts, we provide a structured framework for comprehending the various dimensions of robustness.
- **Robustness metrics.** We summarize the metrics commonly employed to measure the robustness of neural networks within

existing evaluation methods and divide them into two categories: local and global metrics. The evaluation of robustness heavily relies on the effective quantification of perturbations. To this end, we thoroughly examine and summarize various image perturbation measurement techniques employed in image recognition. This includes metrics designed for measuring the magnitude of perturbations and representations that depict the perturbation range.

- **Robustness assessment methods.** We review and analyze the verification and testing techniques used for robustness assessment. For verification assessment approach, we investigate the meaning of verification and validation within AI to explain how to measure and evaluate the robustness based on neural network verification, formal verification for lower bounds of AR and statistical verification for probabilistic robustness are involved. The testing assessment approach involves adversarial testing for upper bounds of AR and benchmark testing for CR, as well as the methods for analyzing test adequacy, which can be categorized into neuron coverage metrics and input domain coverage metrics. We highlight their strengths, limitations, and applicability in practical scenarios.

- **Challenges and Future Directions.** We identify several open challenges and potential future research directions in the field of neural network robustness assessment. One crucial aspect is the need for standardized certification processes or the establishment of effective benchmarks for robustness testing. By defining standardized procedures or benchmarks, we can provide a consistent and reliable framework for evaluating the robustness of neural networks, enabling fair comparisons between different approaches and models.

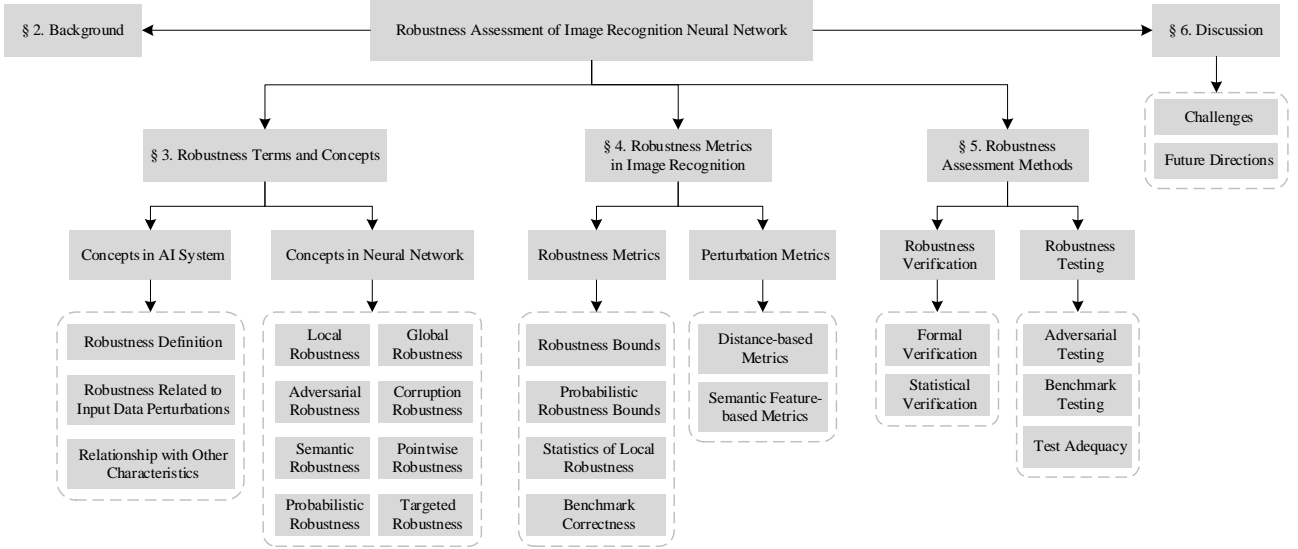

Figure 1: The structure of this survey.

The structure of this survey is shown in Figure 1. In Section 2, we provided the background about the robustness assessment subject in this survey. In Section 3, we review the definition of robustness and related terms for AI systems, discuss the relationship between robustness and other AI quality characteristics, and analyze the concepts and terms proposed in neural network robustness assessment. In Section 4, we summarize existing robustness metrics applicable to image classification models in terms of local robustness and global robustness, and analyze image perturbation metrics used to measure robustness. In Section 5, we categorize the existing robustness assessment methods into verification and testing, and emphasize the strengths and limitations of each method. In Section 6, we identify the challenges and future directions of robustness assessment in image recognition.

## 2 Background

To provide a more effective overview of neural network robustness assessment, we first outline the background of robustness assessment with the aim of clarifying the different research objects of AI robustness, establishing the concepts of system-level robustness and algorithm-level or model-level robustness, and thus better summarizing the terms and concepts used in robustness assessment.

The field of Artificial Intelligence robustness covers a wide range of objects and targets, including AI system [1, 8, 9], AI-based system [10], AI component [10], AI module [4, 11] Machine Learning (ML) systems[12], Neural Networks[13, 14] etc. These concepts represent different product levels, and in general, they can be categorized into two levels: system level and algorithm (or model) level. ISO/IEC DIS 22989:2021[1] defines an AI system as an engineered system featuring AI. ISO/IEC TR 29119-11: 2020[10] gives guidelines on the testing of AI-based systems, and the standard defines AI-based system as system including one or more components implementing AI. Felderer and Ramler [15] provide an overview of quality assurance for AI-based systems, defining an AI-based system as a software-based system that comprises AI components in addition to traditional software components. The standards DIN SPEC 92001-1:2019 [11] and -2:2020 [4] apply to AI modules which is defined as software module consisting of AI algorithms, this concept is equivalent to the AI components. In addition, Zhang et al [12] reviewed the Machine Learning (ML) system testing, and defined an ML system as a system consisting of three components: data, learning program, and framework, similar to the AI component and AI module, it refers to the product level, i.e., a software module that implements AI. The same is the case for ML systems defined in ISO/IEC 23053:2021[9].

Machine learning algorithm (ML algorithm) and machine learning model (ML model) are two common terms. Algorithms and models have different priorities, with algorithms are used to describe the method and process of learning, while models are used to refer to products such as regression vectors or functions, if-then statement trees, weight matrices, etc., that are obtained from the training process [16]. Many standards [1, 8-10] provide clear definitions of ML algorithms and models. In general, an ML algorithm is the process running with data to create an ML model, and an ML model is the output of an ML algorithm running on data.

In the context of trustworthy AI, particularly AI robustness evaluation, the existing research findings and methods primarily performed on models trained with deep neural network algorithms, although in many studies the terms "algorithm" and "model" are used interchangeably without clear distinction. Consequently, the assessment of robustness predominantly occurs at the algorithm (or model) level of AI products. This survey focuses on the DNN classification models for image recognition.

## 3 Robustness Terms and Concepts

In the field of trustworthy AI, the definitions of robustness in current standards and reviews primarily pertain to AI systems, which refers to AI systems combining hardware and software or software subsystems implementing AI algorithms. The definition of robustness for algorithm-level AI models, such as neural network models, aligns with the system-level definition, but there are some specific robustness concepts. In Section 3.1, we provide a concise summary of the existing robustness definitions applicable to AI systems, and in Section 3.2, we conduct an analysis of the robustness terms and concepts relevant to neural network.

**3.1 Concepts in AI System**

*3.1.1 Definition of Robustness*

Robustness is an important property that has been extensively studied in the field of trustworthy AI. There is now a widely accepted definition of robustness, which refers to **the ability of (degree to which) an AI system to maintain its level of performance under any circumstances (including external interference or harsh environmental conditions)** [1, 2, 8, 13]. Such interferences or perturbations may arise in various system components, including data, learning program, and framework. Robustness to changes in input data is of great significance and has received the most attention. The concepts of robustness and security in AI systems are closely related and are often considered equivalent in existing research, particularly in the context of adversarial robustness.

In 2022, ISO/IEC DIS 25059:2022 [2] published, it extends SQuaRE (Systems and Software Quality Requirements and Evaluation) to AI systems and defines the quality model for AI systems based on the system/software quality model [17]. In this updated model, robustness is introduced as a new sub-characteristic of reliability. As defined in this standard, robustness refers to "degree to which an AI system can maintain its level of performance under any circumstances." This definition is derived from ISO/IEC DIS 22989:2021[1]. Notably, the refinement from "ability" to "degree" emphasizes the quantitative aspect of robustness, reinforcing the requirement for a measurable level of robustness. The standard also provides some circumstances of robustness including:

- the presence of unseen, biased, adversarial or invalid data inputs;
- external interference;
- harsh environmental conditions encompassing generalization, resilience, reliability;
- attributes related to the proper operation of the system as intended by its developers.

ISO/IEC TR 24029-1:2021 [13] offers methods and processes for evaluating the robustness of neural networks. The definition of robustness for neural networks, which is the core component within AI systems, follows the definition of system robustness. Since this standard primarily emphasizes the quantitative assessment of neural network robustness, the definition of AI system robustness within the standard "mainly describes data input circumstances such as domain change".

Zhang et al. [12] provided a comprehensive overview of research on ML system testing. As discussed in section 2, an ML system represents a typical AI system, particularly the software subsystem implementing AI functions which consists of data, learning program, and framework. They defined the robustness of ML systems as the resilience to perturbations based on IEEE Std 610.12-1990 [18]. Specifically, it refers to the difference between correctness of perturbed and unperturbed ML systems, with perturbations can be present in any ML components. This definition of robustness is based on the concept of correctness.

In addition, robustness of AI systems is addressed in pertinent standards such as ISO/IEC 23894:2023 [19] and SAE J 2958-2020 [20]. ISO/IEC 23894:2023 emphasizes robustness as an objective to AI risk management, highlighting the importance of maintaining performance under various usage circumstances and handling invalid inputs or stressful environmental conditions. SAE J 2958-2020 specifically discusses the robustness, fault tolerance, and reliability of Unmanned Ground Vehicles (UGVs) in harsh and hostile environments, emphasizing the need for AI systems to exhibit robustness in challenging conditions.

ISO/IEC TR 24028:2020[8] is another standard that provides insights into the concept of robustness in different types of AI systems. For classification tasks, robustness is the ability of the AI system to consistently classify known inputs as well as inputs within a certain range, meaning the system can correctly classify both familiar inputs and unknown inputs that are not significantly different from the known inputs.

### 3.1.2 Robustness Related to Input Data Perturbations

According to the definitions above, AI system robustness encompasses a wide range of anomalies originating from various sources, including data, hardware, and operating environment. This is a broad definition of robustness. Existing researches primarily focus on abnormal conditions related to input data. These abnormalities may include unseen, biased, adversarial, or invalid data [2], as well as domain change of input [13, 14]. This case is a confined meaning of robustness. The robustness associated with abnormal input data is also referred to as out-of-distribution (OOD) [21-23] issue in many studies.

ISO/IEC TR 24028:2020 [8] highlights that robustness involves the capability of a system to handle unknown data and operate effectively in rapidly changing environment. ISO/IEC DIS 22989:2021 [1] and ISO/IEC TR 24029-1:2021 [13] define the terms typical data or typical inputs to refer to known inputs that the AI system has been trained on, robustness is associated with atypical data, one example is domain change. ISO/IEC DIS 24029-2:2022[14] defines domain as "set of possible inputs to a neural network characterized by attributes of the environment", and highlights that the domain reflects the limitations of current AI technology, i.e., neural networks can only achieve their goals on appropriate inputs, and robustness is closely related to the domain in which they operate. Domain change is also called dataset shift in some studies [4, 24].

DIN SPEC 92001-2:2020 [4] introduces two types of robustness: adversarial robustness (AR), which addresses deliberate attacks through carefully crafted harmful inputs, known as adversarial samples, and data corruption robustness (CR), which focuses on handling differences between datasets during development and deployment phases, called data distributional shift or dataset shift. This standard also considers a specific aspect of data corruption robustness, namely spatial robustness, which pertains to the module's ability to handle geometric transformations like translation and rotation.

Some examples of data anomalies or perturbation scenarios considered in existing studies on AI system robustness include:
1) Change in the domain or dataset shift;
2) Unseen data, which refers to data that is not present in the training set；

3) Intentional attack input, such as adversarial instance or adversarial example;

4) Data corruption input, such as noisy data, distorted signals, or geometric transformations applied to images;

5) Invalid input;

6) ……

*3.1.3 Relationship between Robustness and Other Quality Characteristics*

*3.1.3.1 Relationship to Trustworthiness, Resilience and Reliability*

ISO/IEC TR 24028:2020 [8] defines trustworthiness as the ability to meet stakeholders' expectations in a verifiable way and gives some characteristics of trustworthiness, such as reliability, availability, resilience, security, privacy, etc. While not explicitly including the robustness property, it is noted that robustness assurance is a part of AI trustworthiness verification. On the other hand, the standard also points out that robustness encompasses resilience, reliability and potentially more attributes, as related to proper operation of a system as intended by its developers. The same view in ISO/IEC DIS 22989:2021 [1] about the relationship between robustness, trustworthiness, resilience, and reliability. Zhang et al. [12] equate robustness with resilience in the context of machine learning (ML) systems.

According to the AI quality model defined in ISO/IEC DIS 25059:2022 [2], robustness is a sub-characteristic of reliability, alongside with maturity, availability, fault tolerance, and recoverability. Robustness is a newly introduced sub-characteristic specifically for AI systems, while others are defined in the original system/software quality model. Reliability is related to the specified functions, conditions, and operating times. Considering the challenges of specifying the function of an AI system due to its uncertainty and complexity, it is reasonable to consider robustness as a sub-characteristic of reliability assurance. However, it should be noted that ISO/IEC TR 24028:2020 [8] and ISO/IEC DIS 22989:2021 [1] think that robustness should encompass reliability.

In summary, robustness is recognized as an attribute of AI trustworthiness. Trustworthiness is a broader concept, reliability, resilience, etc. are also included, which is consistent with traditional software or systems. However, the relationship between robustness, resilience, and reliability is still not fully understood. But it is clear that robustness plays an important role in ensuring AI reliability because of the complex operating environments and limited training data of models.

*3.1.3.2 Relationship to Fault Tolerance*

The AI quality model [2] considers robustness and fault tolerance as two sub-characteristics of reliability. ISO/IEC 25010:2011 [17] defines fault tolerance as the degree to which a system, product or component operates as intended despite the presence of hardware or software faults. ISO/IEC TR 24028:2020 [8] describes fault tolerance for AI systems as the ability to continue to operate when disruption, faults and failures occur within the system, potentially with degraded capabilities. In the field of software engineering, IEC 62628-2012 [25] refers to fault tolerance as a strategy for software fault control and trustworthiness realization, and is employed during design and implementation to ensure software trustworthiness.

According to the definitions above, fault tolerance primarily deals with internal anomalies within a system, while the current definition of AI system robustness focuses on external disturbances or changes in environmental conditions, particularly regarding changes in data inputs during runtime. So one can distinguish between the two properties in terms of internal and external exceptions.

However, a different view is shown in other studies. Barmer et al. [24] suggested that the robustness of an AI system is associated with model errors and unmodeled phenomena, which are internal exceptions due to the algorithm and training process. The definition of ML system robustness by Zhang et al. [12] supports this viewpoint, considering perturbations in various aspects including data, learning programs, and framework, involving both internal and external exceptions.

Further research is needed to clarify the relationship between robustness and fault tolerance in AI systems. Understanding the function of AI systems and effectively categorizing exceptions and perturbations is essential for drawing conclusive insights. Moreover, it is important to recognize the critical role of data in AI systems, with data being treated as a component within the system. The quality of data during the training phase, including both train set and test set, represents a significant risk resource for AI systems. Therefore, when differentiating between robustness and fault tolerance based on internal and external exceptions, the quality of data used during training and the data encountered during real-world applications are important factors to consider. The data quality during the training phase is closely related to fault tolerance, while robustness primarily focuses on the system's ability to handle external disturbances or changes observed in usage data.

*3.1.3.3 Relationship to Security and Safety*

Barmer et al. [24] conducted a study on robust and secure AI, providing a clear distinction between the concepts of robustness and security in AI systems. Robustness is concerned with the ability to maintain correct outputs for unmodeled parts of the input space, particularly the perturbations due to changes (e.g., noise, sensor degradation, context shifts, etc.) in the operating environment or new environments. They categorized robustness into 1) robustness against model errors and 2) robustness against unmodeled phenomena. Security, on the other hand, addresses hazards from specific threat patterns, such as 1) intentional subversion and 2) forced failure, i.e., perturbations associated with malicious attacks, such as adversarial attacks.

DIN SPEC 92001-2:2020 [4] further divides robustness into adversarial robustness (AR) and corruption robustness (CR). AR deals with scenarios involving active adversaries and requires continuous defense against attacks. CR is caused by natural factors. Regarding their relationship, AR is considered a security concern while CR is viewed as a safety issue.

Zhang et al. [12] defined security of ML systems as resilience against potential harm, danger, or loss made via manipulating or illegally accessing ML components, similar definitions by Barmer et al. [24] and DIN SPEC 92001-2:2020 [4]. They also note that robustness and security are closely related, as low robustness can make a system vulnerable to adversarial attacks or data poisoning. However, security encompassed more than just robustness and include aspects like model stealing or refining.

Cheng et al. [26] introduced four dependability metrics for neural networks, including robustness, interpretability, completeness, and correctness. Robustness addresses various effects such as distortion or adversarial perturbation, and is closely related to security. In neural network robustness assessment, the terms robustness and safety are often used interchangeably, where robustness boundary verification is also referred to as maximum safety radius verification [27-30]. Bu Lei et al. [31] also pointed out that robustness is a prerequisite for security.

In conclusion, while there are attempts to differentiate between robustness and security in standards, in current research and practice, they are often treated as equivalent concepts.

*3.1.3.4 Relationship to Uncertainty*

Chen et al. [32] conducted an empirical study that demonstrated a positive correlation between adversarial robustness and prediction uncertainty. They categorized uncertainty in deep learning into three types: model uncertainty, data uncertainty, and prediction uncertainty. Prediction uncertainty refers to the level of confidence exhibited by a model in its predictions, where lower confidence indicates higher prediction uncertainty. They explained that DNNs with deterministic classification boundaries are more susceptible to adversarial attacks. On the other hand, models with higher prediction uncertainty achieve a better balance by ensuring that the classification boundaries are maximally distant from each class of data, making the attack more difficult. Building upon this insight, robustness of a model can be enhanced by increasing the uncertainty of its predictions, and the accuracy can also be maintained.

*3.1.3.5 Relationship to Functional Correctness*

ISO/IEC DIS 25059:2022 [2] introduces a new sub-characteristic called functional adaptability in the AI quality model and modifies the definition of functional correctness. It is often difficult for AI systems to provide functional correctness because a certain error rate is allowed in their output. Functional adaptability refers to the ability of the system to adapt itself to a changing environment it is deployed in by learning from new training data, operational input data, and previous actions. As can be seen from the definitions, AI system robustness is also concerned with situations such as runtime input data and the results of previous behaviors and can reflect performance changes in these situations, functional adaptability highlights learning from these situations.

Zhang et al. [12] provide definitions for correctness and robustness in ML systems. Correctness is defined as the probability that the ML system "gets things right," and it is closely related to functional correctness in ISO/IEC DIS 25059:2022. Metrics like accuracy, precision, recall, and AUC are commonly used to measure correctness. Robustness is defined on the basis of correctness, expressed as the difference in correctness between the system with and without perturbations, i.e., the ability of the system to maintain correctness in the presence of perturbation.

ISO/IEC DIS 25059:2022 [2] also highlights the trade-offs among functional correctness, robustness, and performance efficiency. Chen et al. [32] studied the accuracy-robustness trade-off in adversarial training and proposed a method to alleviate it by maximizing prediction uncertainty. Zhang et al. [33] explored the theoretical principles underlying the trade-off between adversarial robustness and accuracy. The trade-off between accuracy and robustness is further discussed by Li et al [34], examining its evolution and inherent/non-inherent existence.

However, Yang et al. [35] challenged the inherent trade-off between robustness and correctness. They argued that this trade-off can be avoided and provided an explanation based on the local Lipschitz function. They conducted experiments combining dropout with robust training to get better generalization thus avoiding the decrease in test accuracy, disproving the belief that a trade-off between accuracy and robustness is inevitable.

*3.1.3.6 Relationship to Explainability*

ISO/IEC 23894:2023 [19] discusses the relationship between explainability and robustness of AI systems. Explainability is defined as the property of an AI system that the important factors influencing a decision can be expressed in a way that humans can understand. Since the behavior of DNNs is difficult to understand, it is important to effectively express these factors to understand why the AI system makes certain decisions and whether it can make the right decision in all situations. The uncertainty and lack of explainability of AI behavior lead to risks that will affect many attributes of AI, such as robustness. The inherent difficulty in explaining the nonlinear nature of neural networks always leads to unexpected behavior, making it difficult to evaluate and analyze their robustness.

**3.2 Concepts in Neural Network**

The neural network model is a key component of AI systems, and the definition of its robustness follows the definition for AI system. However, analyzing the robustness of neural networks is challenging due to their complexity, which includes nonlinearity, nonconvexity, and limited interpretability. Researchers have dedicated considerable effort to studying the robustness of neural networks. As a result, a wide range of research findings has emerged, covering different concepts that describe or quantify the robustness of neural networks from multiple perspectives. To maintain symbol consistency throughout this survey, symbols referenced from various papers have been adjusted. Some key symbols are listed in Table 1.

Table 1: Symbols used in this survey.

| Symbol | Description |
|---|---|
| $F$ | Neural network model being assessed. |
| $x$ | Origin/clean image, a single point to be assessed. |
| $x'$ | Perturbed image of $x$. |
| $\|\cdot\|$ | Distance measure between two images. |
| $c$ | Classification result of $F$ for input $x$. |
| $F(x)$ | The probability of class $c$. |
| $\delta$ | The given degree of attacks, corruptions, or perturbations. |
| $\epsilon$ | Maximum variation can be tolerated at the output in robustness verification. |
| $\rho$ | Local robustness measurement. |
| $X$ | Origin/clean dataset. |
| $T$ | Robustness test suite. |

*3.2.1 Local Robustness and Global Robustness*

According to ISO/IEC DIS 24029-2:2022 [14], robustness properties can be categorized as either local or global. Local robustness refers to the model's ability to maintain its output within specific regions in the input space, while global robustness examines or measures the model's overall robustness across the entire input space.

*3.2.1.1 Local Robustness*

The present definitions and studies of local robustness primarily focused on a given test input. Specifically, local robustness is defined within the neighborhood of a sample, which refers to a range of variations or perturbations near the sample. To illustrate, let's consider an image correctly classified as a car. In this case, the local robustness can be defined such that any image generated by rotating the original image within a range of 5 degrees should also be classified as a car.

In the work by Leino et al. [36], local robustness is defined as: a neural network model $F$ is said to be $\delta$-locally-robust at point x if it makes the same prediction on all points in the $\delta$-ball centered at x, which can be described as follows:

**Local Robustness**: If a neural network model $F$ is $\delta$-locally-robust at a given point x with respect to the distance metric $\|\cdot\|$, it implies that for $\forall x'$, the following condition holds true:

$$\|x - x'\| \leq \delta \rightarrow F(x) = F(x')$$

The same definition can also be found in the study conducted by Katz et al. [27]. They describe local robustness in the context of classification models and infinite norm. Specifically, a network is $\delta$-locally-robust at input point x if for every $x'$ such that $\|x - x'\|_\infty \leq \delta$, the network assigns the same label to x and $x'$.

Zhang et al. [12] introduced the concept of "Local Adversarial Robustness" within the context of machine learning systems. This definition diverges from the definitions presented in the aforementioned papers by explicitly characterizing $x'$ as the test input generated by an adversarial perturbation to x. By incorporating this notion, the definition of local robustness encompasses local adversarial robustness as a distinct case that elucidates the manner or origin of $x'$ generation.

It is obvious that local robustness primarily pertains to the robustness around an input point. Many researches on evaluating neural network robustness has concentrated on local robustness.

*3.2.1.2 Global Robustness*

There is a lack of an agreed definition of global robustness and the meaning varies across different studies. Effective methods for analyzing, evaluating, and quantifying global robustness remain scarce. The following are some existing definitions of global robustness.

Katz et al. [27] define local robustness as the measure for a specific input x, while global robustness applies to all inputs simultaneously. They analyze global robustness by comparing the outputs of input x and its perturbed inputs $x'$ on two replicas of DNN, $N_1$ and $N_2$. They describe global robustness as follows: Let $x_1$ and $x_2$ denote separate input variables for $N_1$ and $N_2$, respectively, and $x_2$ represents an adversarial perturbation of $x_1$ that satisfies $\|x_1 - x_2\|_\infty \leq \delta$. $N_1$ and $N_2$ assign output values $p_1$ and $p_2$ respectively. If $|p_1 - p_2| \leq \epsilon$ holds for every output, the network is $\epsilon$-globally-robust. Zhang et al.[12] define global robustness as:

**Global Robustness 1**: A neural network model $F$ is $\epsilon$-globally-robust if for $\forall x, x'$,

$$\|x - x'\|_p \leq \delta \rightarrow F(x) - F(x') \leq \epsilon$$

Leino et al. [36] argue that it is impossible to achieve local robustness at every single point simultaneously. Unless the model is entirely degenerate, there will always exist a point arbitrarily close to the decision boundary. To address this challenge, they propose the concept of global robustness by introducing an additional class denoted as "⊥", which signifies the classifier's refusal to provide a prediction and indicates a violation action. When a point is classified as ⊥, it implies that it cannot be certified as globally robust. By introducing the ⊥ class, the classifier ensures the existence of at least one partition of width $\delta$ between any pair of regions with different predicted labels to be assigned as ⊥. In other words, no two points within a distance of $\delta$ would be labeled with different non-⊥ classes. To define the relation "⊥=" based on the ⊥ class, they denote $c_1 \perp= c_2$ as $c_1 =\perp$ or $c_2 =\perp$ or $c_1 = c_2$. This definition enables the establishment of global robustness, and the inclusion of the ⊥ class allows for the training of a classification model that achieves global robustness.

**Global Robustness 2**: A neural network model $F$ is $\delta$-globally-robust with respect to the norm $\|\cdot\|$ if for $\forall x_1, x_2$ satisfies:

$$\|x_1 - x_2\| \leq \delta \rightarrow F(x_1) \perp = F(x_2)$$

Kabaha et al. [37] also presented a concept of minimum globally robust bound, it is proposed for classes of classification model, not the input points. This global robustness is used to ensure that the output of each class is invariant to perturbations.

In addition to the definitions above, the robustness measured by the statistical characterization of local robustness radius, as discussed in [28, 38, 39], is also the global robustness. It is a profile of the local robustness of multiple test inputs. Besides, it is important to note that the robustness from benchmarking methods [40, 41] which describes the overall robustness level of a model by evaluating its performance on a benchmark specifically designed for robustness testing is differs from the definition of global robustness in robustness verification. A recent study by Zhang et al. [42] proposed DeepGlobal, a framework for global robustness verification. This framework goes beyond the limitations of the test set data samples and is capable of identifying all potential adversarial hazardous regions of a neural network. The definition of global robustness in the study is based on spatial region division.

In conclusion, further investigation is required to establish a precise and accurate definition of global robustness for neural networks, considering the interrelationships among the various existing definitions and concepts.

*3.2.2 Adversarial Robustness, Corruption Robustness and Semantic Robustness*

Adversarial robustness is a widely studied concept in neural network robustness. In many cases, neural network robustness assessment is conducted on adversarial examples generated by adversarial attack algorithms, and this evaluation result is called adversarial robustness [12]. It is generally accepted that the robustness against adversarial perturbations is closely related to their security [4, 12, 27-30]. The defense techniques [43-50] for enhancing adversarial robustness is also an important research topic.

The concept of corruption robustness (CR) is defined in DIN SPEC 92001-2:2020 [4] and contrasted with adversarial robustness (AR). AR refers to the ability of an AI module to cope with adversarial examples (or adversarial perturbations), while CR focuses on noisy signals or changes in the underlying data distribution. Within CR, there is a subcategory called spatial robustness, which pertains to robustness against geometric transformations like translation and rotation. These two concepts describe and categorize robustness in terms of the different ways in which data perturbations can be generated. AR focuses on carefully crafted adversarial inputs, considering an active adversary in an ongoing "arms race" between attacks and defense strategies. On the other hand, CR addresses issues arising from natural factors such as hardware deterioration or shifts in data distribution. Both AR and CR aim to ensure the model performance on inputs that are unlikely or out of the distribution of training data. Adversarial examples can be seen as an intentionally designed change to degrade the performance. While CR is optimization-free because there is no intentional design.

The term "semantic perturbation" has led to the concept of semantic robustness. Huang et al. [51] explored four semantic perturbations that significantly affect the performance of image recognition systems, including rotation, scaling, cropping, and tilting and suggested that it is not necessary to verify the complete robustness of neural networks against small perturbations such as adversarial perturbations under the condition of limited resources, but focus on the robustness against specific perturbations under specific risk scenarios. Similarly, Mohapatra et al. [52] studied the verification of semantic robustness against the change on brightness, lightness, contrast, and rotation. Studies on semantic robustness shares similarities with CR but is not equivalent. Researchers added small perturbations on semantic features to analyze semantic robustness, referring to these perturbations as "semantic adversarial attacks". It mainly focuses on small perturbations on semantic feature rather than pixel modifications, which is different from traditional pixel-based adversarial attacks (such as FGSM, C&W, etc.), as studied in papers [53-58]. Some works also verified and estimated the semantic robustness bounds on semantic adversarial attacks and semantic feature-based metrics [53].

Overall, these concepts of robustness are defined based on different perturbation sources or generation methods and are used to evaluate the resilience of neural networks against specific perturbations.

*3.2.3 Pointwise Robustness*

Bastani et al. [39] introduced the concept of "pointwise robustness". It used to determine the minimum perturbation needed to change the output of a neural network in the neighborhood of a given input point x, i.e., the closest adversarial samples to that point [3]. This adversarial perturbation refers to a small alteration in the input that does not significantly impact human judgment but does affect the predicted output of the neural network. Therefore, pointwise robustness is an approximation of local robustness bound, more precisely, an upper bound. Pointwise robustness is based on the definition of local robustness (denoted as $(x, \delta)$-robust in the paper), as follows:

$(x, \delta)$-**Robust**: A neural network model $F$ is robust at x if a small perturbation $\delta$ to x does not affect the assigned label. A neural network model $F$ is considered $(x, \delta)$-robust if for $\forall x'$ satisfies:
$$\|x' - x\|_\infty \leq \delta, F(x') = F(x)$$

**Pointwise Robustness**: The pointwise robustness $\rho(F, x)$ of a neural network model $F$ at a specific input point x is defined as the minimum value of $\delta$ for which F fails to satisfy the $(x, \delta)$-robust property, i.e.:

$\rho(F, x) \stackrel{\text{def}}{=} \inf[\delta \geq 0 | F$ is not $(x, \delta)$-robust]

The concept of pointwise robustness emerges from a broader emphasis on determining the performance boundaries of a neural network in the context of robustness assessment. Adversarial attack algorithms and testing techniques can establish a lower bound on the magnitude of input perturbations that violate the robustness condition, while definition of the robustness boundary is formulated as an upper bound, indicating that any perturbation beyond this boundary range will lead to incorrect model outputs.

*3.2.4 Robustness Bounds*

Many researches have investigated the maximum robustness range of neural networks. This involves analyzing the maximum range of input perturbations that a model can tolerate while still producing correct outputs. Formal verification methods [27, 59-61], such as satisfiability modulo theory (SMT) and mixed-integer linear programming (MILP), are used to determine the robustness bound. Additionally, there are various approximation methods [62-66] to verify the robustness range, they provide an approximate lower bound. These methods aim to ensure that the model remains resilient to perturbations within this boundary but may not provide guarantees against perturbations outside the boundary.

Ji et al.[67] highlighted that the robustness boundary is for a specific sample, represents the maximum perturbation range that the sample can tolerate while still ensuring correct predictions by the model. In other words, within this boundary, the model's classification decision for the given sample remains unchanged. It can be seen that the robustness boundary is also a local robustness definition. They also provided a definition of the robustness boundary for deep learning classification models as follows:

**Robustness Bound**: let the dimension of the input sample x be $d$, the number of output categories be $K$, the neural network model be $F: \mathbb{R}^d \to \mathbb{R}^K$, and the categories of the input sample be $c = argmaxF_j(x), j = 1,2,\ldots,K$. Under the assumption of the $l_p$ space, the model provides the $\delta$-robustness guarantee to indicate that the model's classification decision for x does not vary within $\delta$-size around this sample $l_p$-space.

Denoting the exact bound and lower bound of the maximum robustness range differs from the upper bound defined by pointwise robustness in section 3.2.3 [31, 67], and pointwise robustness can also be considered as a robustness bound. Differences and connections between the two concepts can be understood as follows:

1) Robustness exact bounds and approximate lower bounds refer to the maximum provable robustness of a neural network. In contrast, pointwise robustness is an approach based on testing that can only provide guarantees that the model is not robust for perturbations exceeding a certain threshold.

2) Pointwise robustness approximates the critical value from the region where the region where the adversarial attack algorithm is not robust. On the other hand, approximate verification of the maximum robust range aims to approximate the critical value from the region where the model remains robust. Therefore, pointwise robustness, based on adversarial attacks, represents the minimum attack radius rather than the maximum robust radius.

3) The lower bound of robustness guarantees that the model remains robust within a specific region, while the upper bound of

robustness indicates that the model becomes non-robust outside that region. When it is not possible to determine the exact bound, the lower and upper bounds of robustness can provide bilateral guarantees for the robustness of the neural network.

*3.2.5 Probabilistic Robustness*

Probabilistic robustness quantifies the level of robustness using a statistical approach. The concept was introduced by Mangal et al. [68], they argued that the traditional definition of robustness against adversarial inputs is a worst-case analysis and unlikely to be satisfied and verified by neural networks, so a probabilistic description of robustness is proposed. Unlike adversarial robustness, probabilistic robustness does not require the nearest counterexample of the input point. Instead, it calculates the probability that the model remains correct behavior within an acceptable range of perturbations. Their definition is as follows:

**Probabilistic Robustness 1**: Let F represent the function represented by the neural network. F(x) denotes the output produced by the neural network for a given input x. Let $x$ and $x'$ be pairs of real inputs with a distance not exceeding $\delta$. The symbol $\|\cdot\|$ represents the distance metrics on the input and output spaces. D denotes the input distribution, k is a Lie constant, and $\epsilon$ represents the specified robustness condition. If the following inequality holds:

$$P_{x,x'\sim D}(\|F(x') - F(x)\| \leq k * \|x' - x\| \mid \|x' - x\| \leq \delta) \geq 1 - \epsilon$$

Then F is considered to satisfy probabilistic robustness, and the robustness probability is $\epsilon$.

Webb et al. [69] also defined probabilistic robustness and explored approach to estimate it through statistical verification. Their definitions are presented below (with appropriate notation modifications for consistency within this paper):

**Probabilistic Robustness 2**: The neural network F produces its final output through the softmax layer function, denoted as $F_\theta(x) = \text{softmax}(z(x))$. An input $x'$ is considered an adversarial sample if for $x' = x + \delta$, $\text{argmax}_i z(x')_i \neq \text{argmax}_i z(x)_i$. Consequently, we define the difference function $s(x') = max_{i \neq c}(z(x')_i - z(x')_c)$, where $c$ represents the label of $x$. When $s(x') \geq 0$, it indicates that the input is an adversarial sample (perturbed in a way that can be addressed by the method). To quantify the probability of model failure, we integrate over the probability of $s(x') \geq 0$ with respect to the assumed input distribution. Formally, this can be expressed as follows:

$$[p,s] \triangleq P_{X\sim p(\cdot)}(s(x') \geq 0) = \int_X \mathbb{I}[s(x') \geq 0]p(x)dx$$

Probabilistic robustness provides a broader definition of robustness. However, the accuracy of estimation depends on two factors: proper assumptions regarding the distribution of perturbations and the sufficient test cases.

*3.2.6 Targeted Robustness*

Targeted robustness is a concept specifically defined for classification tasks, focusing on the robustness towards specific target labels or categories. Gopinath et al. [70] introduced this notion to ensure that a classification model does not assign a given input to a given target label. The neural network attacks in the NIPS competition organized by Kaggle is also divided into two categories, targeted attacks aim to generate adversarial samples that cause the classifier to misclassify the input into a specific class [71], as opposed to non-targeted attacks that aim for misclassification without specifying a particular target class [72]. DIN SPEC 92001-2:2020[4] also defines targeted attacks as those where the adversary tries to produce inputs that force the output of the classification model to be a specific target class. Another relevant concept is top-k prediction robustness, which refers to the ability of a model to ensure that the correct category remains within the top-k predictions for perturbed inputs [73].

Targeted robustness is a concept that focuses on the resilience of a neural network to specific target labels or categories, rather than guaranteeing robustness against all other labels, as stated by Gopinath et al. that it avoids classifying as a second likelihood label, adversarial samples can also lead to misclassification as other labels [70]. Nonetheless, targeted robustness is still valuable in practice as it relaxes the robustness condition in analysis and evaluation, making it more applicable in scenarios with resource limitations or top-k prediction tasks.

# 4 Metrics for Robustness Assessment in Image Recognition

The assessment of neural network robustness relies on various metrics, which are contingent upon the evaluation methods employed. These metrics offer valuable insights into the robustness of a model from different perspectives. Additionally, different types of task models, such as regression and classification models, utilize distinct metrics. This section summarizes the robustness metrics of neural network models trained for image classification tasks. Metrics used for assessing robustness can be categorized into either local or global, depending on whether they evaluate partial area of the input space or the entire space. Section 4.1 provides an overview of the robustness metrics, and Table 2 shows these metrics along with their respective assessment methods. The robustness is measured primarily by the degree of input perturbation, and the image perturbation metrics are analyzed in Section 4.2.

Table 2 Robustness Metrics and assessment methods for neural networks in image recognition.

| | Metrics | Assessment methods | Techniques |
|---|---|---|---|
| Local Robustness | Exact bound | Complete verification, see 5.1.2.1. | SMT/SAT、MILP |
| | Lower bound | Incomplete verification, see 5.1.2.2. | Convex Relaxation, Abstract Interpretation, Lipschitz Constant, Randomized Smoothing, Interval Boundary Propagation, Cybernetics |
| | Upper bound | Adversarial testing, see 5.2.1. Incomplete verification (upper bound is not the concern of this method). | Adversarial Attack Algorithms: FGSM, DeepFool, C&W, Houdini, BIM, PGD, etc. |
| | Probabilistic bound | Statistical verification, see 5.1.3. | PROVEN、PROVERO、DeepPAC, etc. |
| Global Robustness | Statistics of local radii | Statistics of local robustness, see 4.1.2.1. | Mean, Expectation, Threshold-based Statistics |
| | Benchmark correctness | Benchmark testing, see 5.2.2. | Perturbation generation techniques |

## 4.1 Robustness Metrics

Robustness verification and adversarial testing provide measures such as robustness bounds and probabilities for classification models. These metrics assess local robustness. While benchmarking methods aim to evaluate model robustness by its performance on a designed perturbed test set, which is the global robustness. Detailed definitions for local and global robustness see Section 3.2.1.

*4.1.1 Local Robustness Metrics*

There are two types of local robustness metrics, robustness bound and robustness probability. The robustness bounds are further divided into three types: exact bound, lower bound, and upper bound.

*4.1.1.1 Robustness Bounds*

The robustness bound is a local robustness metric resulting from neural network verifications. It is defined as the verified hypersphere within the input space, where the maximum safe radius represents the robustness bound for a given input point. Typically, the radius of this bound is utilized to quantify the level of robustness. Given the existence of both exact and approximate verification methods, exact bound and approximate lower bound of the robustness bounds can be determined, respectively. Moreover, adversarial testing is employed to compute the exact lower bound of input domain perturbations that induce erroneous model outputs, such as pointwise robustness (Section 3.2.3). Consequently, the radius of the upper bound of local robustness can be expressed by the minimum adversarial distance of this exact bound. There are three existing local robustness bounds metrics: exact bound, lower bound, and upper bound, and their relationships are elucidated in Section 3.2.4. The definitions of the three local robustness bounds are as follows:

**Exact Bound Radius** [62]: Let $c = \mathrm{argmax}_i F_i(x)$ denote the output of the neural network model $F$ for input x, the radius of the exact bound is defined as the smallest perturbation, also referred to as the minimum adversarial distortion in some studies, such that

$$\mathrm{argmax}_i F_i(x + \delta) \neq c$$

The exact bound radius is $\rho_{min} = \|\delta\|_p$.

**Lower Bound Radius** [62]: The lower bound radius $\rho_{cert}$ of the certified robustness satisfies the following two conditions: 1) $\rho_{cert} < \rho_{min}$; 2) $\forall \|\delta\|_p \leq \rho_{cert}$ ($\delta \in \mathbb{R}^d$) such that $\text{argmax}_i F_i(x + \delta) = c$.

**Upper Bound Radius**: can be expressed by the pointwise robustness [39] $\rho(F, x)$, which is the minimum value of perturbation degree generated by the adversarial attack algorithms

$$\rho_{upper} = \rho(F, x) \stackrel{\text{def}}{=} \inf[\delta \geq 0 | F \text{ is not } (x, \delta)\text{-robust}]$$

It also satisfies the following two conditions 1) $\rho_{upper} > \rho_{min}$; 2) $\forall \|\delta\|_p \geq \rho_{upper}$ ($\delta \in \mathbb{R}^d$) such that $\text{argmax}_i F_i(x + \delta) \neq c$.

The exact robustness bound can be approximated from its upper and lower bounds.

*4.1.1.2 Probabilistic Robustness Bounds*

Probabilistic robustness, derived from statistical verification, introduces a probability-based metric to assess robustness within the δ-neighborhood input domain of a neural network model for a given input point. This metric relaxes the requirements of formal robustness verification and eliminates the need to find the nearest counterexample, thereby accommodating non-adversarial scenarios. Instead of ensuring robustness for all samples, probabilistic robustness focuses on ensuring robust for most samples within the verification region by quantifying the probability of inputs satisfying the robustness condition. It can be defined for both local and global robustness, see 3.2.5.

The statistical verification-based approach provides a measurement of local probabilistic robustness, which includes the verification region and the probability within that region. This region can be interpreted as a confidence boundary for local robustness. By converting the integral of probabilistic robustness definition in section 3.2.5 into a statistical estimate of sampling within the input distribution, the following formula for the probabilistic robustness metric can be obtained:

**Probabilistic robustness** [69]: $\hat{P}_{X \sim p(\cdot)}(s(X) \geq 0) = \frac{1}{N} \sum_{n=1}^{N} \mathbb{I}[s(x_n) \geq 0]$, $x_n \stackrel{i.i.d}{\sim} p(\cdot)$。

Existing methods for statistical verification of local probabilistic robustness and estimation of their robustness probability measures are analyzed in Section 5.1.3.

*4.1.1.3 Relationships among metrics*

Figure 2 illustrates the relationship between the existing local robustness bounds and the sample prediction results, indicating whether the neural network produces correct prediction outputs. It also showcases the relative positions of the exact, lower, and upper bounds, along with the confidence boundary.

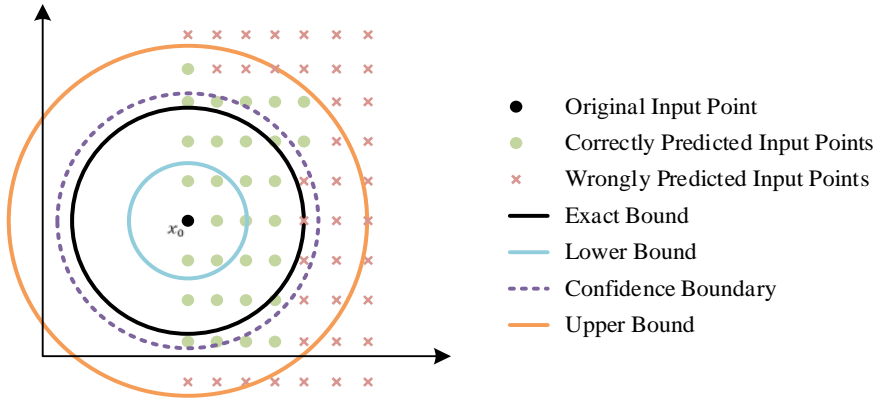

Figure 2: Schematic representation of the existing local robustness bounds (in 2D input space)

*4.1.2 Global Robustness Metrics*

There are two types of global robustness metrics: metrics based on local robustness statistics, and metrics based on benchmark correctness.

*4.1.2.1 Metrics Based on Local Robustness Statistics*

The global robustness can be estimated by the statistics of local robustness measurements of multiple test input points. Most commonly used local robustness metrics are $l_p$-norms, specifically the radius of a hypersphere. Consequently, global robustness is also measured using statistics of these radii, existing metrics such as the mean, minimum, expectation, and threshold-based statistics, as follows:

1) **Average of local bound radius** [38]

The robustness of the model on a test dataset is calculated as the average ratio of the minimum perturbation value corresponding to each input on the test dataset to the input size, as follows:

$$\hat{\rho}_{\text{adv}}(f) = \frac{1}{|T|} \sum_{x \in T} \frac{\|\hat{r}(x)\|_2}{\|x\|_2}$$

where T is the test dataset and $\hat{r}(x)$ denotes the minimum perturbation value.

2) **Expectation of local bound radius** [28]

Ruan et al.[28] proposed an upper bound estimation method for local robustness based on adversarial testing. They defined global robustness as the expectation of the maximum safe radius on the test dataset, i.e., the expectation of the local upper bound radius. This metric is estimated based on a given test dataset T, which comprises inputs that are independently and identically distributed according to the input distribution of the problem to be solved. Given a neural network model $F$, a bounded input $T_0$ and a distance metric $\|\cdot\|_D$, the global robustness is calculated as follows:

$$\|T - T_0\|_D \stackrel{\text{def}}{=} E_{x_0 \in T_0}(\|x - x_0\|_D) = \frac{1}{|T_0|} \sum_{x_0 \in T_0} \|x - x_0\|_D \ (x_0 \ i.i.d. \ \text{in} \ T_0)$$

3) **Adversarial Frequency and Adversarial Severity** [39]

Bastani et al. [39] introduced two parameters for measuring global robustness based on their definition of pointwise robustness: adversarial frequency and adversarial severity. They are defined as follows:

**Adversarial Frequency**: This metric quantifies the frequency of occurrence of adversarial samples under a given perturbation threshold $\delta$. It is expressed as follows:

$$\emptyset(F, \delta) \stackrel{\text{def}}{=} P_{x \sim D}[\rho(F, x) \leq \delta]$$

For a given data set $X$, if the samples in $X$ are independently and identically distributed, then

$$\hat{\emptyset}(F, \delta, X) \stackrel{\text{def}}{=} \frac{|\{x \in X | \rho(F, x) \leq \delta\}|}{|X|}$$

A higher frequency of adversarial samples indicates that the model is less robust on a larger number of inputs.

**Adversarial Severity**: This metric measures the severity of the adversarial samples under a given perturbation threshold $\delta$, i.e., the average degree of the adversarial attacks, as follows:

$$\mu(F, \delta) \stackrel{\text{def}}{=} E_{x \sim D}[\rho(F, x) | \rho(F, x) \leq \delta]$$

For a given set $X$ drawn i.i.d. from $D$, then

$$\hat{\mu}(F, \delta, X) \stackrel{\text{def}}{=} \frac{\sum_{x \in X} \rho(F, x) \amalg [\rho(F, x) \leq \delta]}{|\{x \in X | \rho(F, x) \leq \delta\}|}$$

A smaller value of adversarial severity indicates that the model is less robust.

The paper also discusses the significance of two parameters. When a model exhibits a high adversarial frequency but a low adversarial severity, it suggests that most adversarial samples are about $\delta$ distance away from the original point $x$. Conversely, when a model demonstrates a low adversarial frequency but a high adversarial severity, it indicates that the model remains robust for the most part, but occasionally severely fails to be robust. Therefore, the adversarial frequency emerges as the more critical metric, as models with low adversarial frequencies tend to maintain robustness in most cases. Adversarial severity, however, serves as a useful tool for distinguishing between models that exhibit similar adversarial frequencies.

*4.1.2.2 Metrics Based on Benchmark Correctness*

The metrics based on benchmark correctness are robustness metrics provided by the robustness benchmark test evaluation method. For classification task models, commonly used correctness metrics include the confusion matrix, accuracy, precision, recall, F-measure, ROC curve, and AUC, among others. ISO/IEC TR 24029-1: 2021 [13] also outlines metrics for evaluating robustness using statistical methods, as depicted in Figure 3.

|  |  | True condition | | Prevalence | Accuracy |
|---|---|---|---|---|---|
|  | Total population | Condition positive (CP) | Condition negative (CN) | $\frac{N_{C+}}{P_{tot}}$ | $\frac{N_{T+}+N_{T-}}{P_{tot}}$ |
| Predicted condition | Prediction positive | True positive (TP) | False positive (FP) | Positive predictive value ($V_{P+}$) Precision, relevance | False discovery rate |
|  |  |  | Type I error | $\frac{N_{T+}}{N_{P+}}$ | $\frac{N_{F+}}{N_{P+}}$ |
|  | Prediction negative | False negative (FN) | True negative (TN) | False omission rate | Negative predictive value Separation ability |
|  |  | Type II error |  | $\frac{N_{F-}}{N_{P-}}$ | $\frac{N_{T-}}{N_{P-}}$ |
|  |  | True positive rate ($R_{T+}$) Sensitivity, recall Probability of detection | False positive rate ($R_{F+}$) Fall-out, probability false alarm | Positive likelihood ratio ($R_{L+}$) |  |
|  |  |  |  |  | Diagnostic odds rate | $F_1$ score |
|  |  | $\frac{N_{T+}}{N_{C+}}$ | $\frac{N_{F+}}{N_{C-}}$ | $\frac{R_{T+}}{R_{F+}}$ | $\frac{R_{L+}}{R_{L-}}$ | $\left(\frac{R_{T+}^{-1}+V_{P+}^{-1}}{2}\right)^{-1}$ |
|  |  | False negative rate ($R_{F-}$) Miss rate | True negative rate ($R_{T-}$) Specificity, selectivity | Negative likelihood ratio ($R_{L-}$) |  |
|  |  | $\frac{N_{F-}}{N_{C+}}$ | $\frac{N_{T-}}{N_{C-}}$ | $\frac{R_{F-}}{R_{T-}}$ |  |

Figure 3: Robustness metrics given in ISO/IEC TR 24029-1 [13].

**4.2 Perturbation Metrics and Perturbation Range**

Robustness measurements are based on quantifying the degree of perturbation. The smallest perturbation that leads to a violation of the robustness condition serves as a measure of the robustness boundaries. The most common metric for perturbation degree is the distance between two images, typically represented by the $l_p$-norm, and the robustness region in input space is modeled as a hypersphere with a radius of $l_p$ distance. In probabilistic robustness verification, the same hypersphere region is employed to sample and estimate robustness probability. Semantic adversarial robustness measures perturbation by comparing semantic features between images, such as rotation angle or brightness changes [52]. Since semantic features are scalar values without direction, there is no need to represent perturbation range and bounds for semantic adversarial robustness in the same manner as input space distance.

*4.2.1 Distance-based Metrics in Input Space and Hypersphere Perturbation Range*

The $l_p$-norm is the most common distance metric used in existing methods, which includes $l_0$、$l_1$、$l_2$ and $l_\infty$. Robustness verification methods are usually proposed for one $l_p$-norm and then extended to others. For instance, Weng et al.[74] proposed a probabilistic robustness verification method for $l_\infty$ and demonstrated that is can be extended to $l_1$ and $l_2$ norms by norm inequalities.

Wicker et al.[75] introduced a feature-guided black-box safety testing method for neural networks, employing the SIFT technique[76] to extract features. These features are manipulated to minimize the $l_p$-norm distance of the adversarial samples, aiming to obtain the global minimum adversarial image.

Croce and Hein[77] investigated verification methods for ReLU networks that offer provable robustness guarantees for hypersphere represented by any $l_p(p > 1)$. They extended the method based on $l_2$ and $l_\infty$ to encompass any $l_p$. The paper demonstrated that $l_2$ and $l_\infty$ guarantees for robustness bounds are sufficient to derive meaningful robustness proofs for all $l_p$ norms. Furthermore, it established that this guarantee is independent of the size of the input space.

Ji et al.[67] conducted an analysis on the effectiveness of each $l_p$ when utilized for model defense. They concluded that defending against attacks at one $l_p$ distance proves ineffective against attacks launched at another norm. For instance, the $l_0$ robustness of $l_\infty$ defense is even lower than that of a network without defense, and it remains susceptible to $l_2$ perturbations. Thus, relying solely on a single norm distance metric for robustness assessment is inadequate.

Carlin and Wagner[78] highlighted the importance of evaluating neural networks with defenses for robustness using robust anti-attack algorithms. They emphasized the necessity of ensuring that the model remains robust against $l_2$ distance metrics, as defenses against $l_2$ attacks can also provide protection against attacks using other distance metrics.

*4.2.2 Semantic Feature-based Metrics and Perturbation Range*

Xiao et al.[79] demonstrated that a neural network model's robustness to $l_p$-bounded adversarial perturbations does not necessarily imply robustness to adversarial spatial transformations. Hence, many researchers have proposed utilizing spatial transformation features of the perturbation or, more generally, semantic features to measure the perturbation and characterize the perturbation range.

Balunovic et al.[80] investigated the robustness of computer vision neural networks to natural geometric transformations, such as rotation and scaling. They suggested that the perturbation metric based on $l_p$-norm is not suitable for geometric transformations of an image. Therefore, they proposed using parametric metrics of geometric transformations to measure perturbation and assess robustness.

Mohapatra et al.[52] pointed out that the $l_p$-norm commonly used in existing verification methods for input samples is not capable of robustly verifying semantic adversarial attacks, such as color shifts and illumination adjustments. As an alternative, they proposed the use of semantic metrics to measure perturbations. These metrics are categorized into two groups: Discrete parameterized perturbations, including translations and occlusions. Continuous parameterized perturbations, such as color shifts, luminance adjustments, contrast variations, and spatial transformations (e.g., rotation). For each category, they suggested specific measure parameters.

Liu et al.[56] introduced a physically-based differentiable renderer, which enables the propagation of pixel-level gradients into the parameter space of luminance and geometry.

Huang et al.[51] proposed utilizing a perturbation parameter metric to gauge the robustness of a neural network against a specific perturbation. They characterize the range of the perturbation by the parameter change interval. For example, in the case of image rotation, the perturbation boundary is denoted as a rotational boundary [0°, 360°], and the distance metric measures the difference in rotational degrees between the original and perturbed images.

Hamdi et al.[53] verified the semantic robustness of neural networks within semantic space by evaluating the semantic distance of the perturbation.

Based on the current state of perturbation metrics in existing semantic robustness evaluations, there is no need to address the issue of modeling and characterizing the range of perturbations. This is because semantic features or semantic perturbation parameters are generally scalar and do not require consideration of directionality.

**5 Robustness Assessment Methods in Image Recognition**

In this section, we review the techniques for assessing the robustness of neural networks. We categorize existing methods into two main groups: robustness verification assessment (see 5.1) and robustness testing assessment (see 5.2). Robustness verification techniques can be used to evaluate the local robustness and offer a verified or certified bound of perturbation around one point.

Robustness testing techniques can achieve global assessment by generating test suite that covers multiple perturbation types and levels from the original clean test set. Verification is suitable only for small-scale neural networks, so testing and statistical approaches are preferred in neural network robustness assessment.

**5.1 Robustness Verification Assessment**

*5.1.1 Overview*

Verification and validation are two important activities in system engineering, and in the domain of AI they have the same meaning [1, 81]. DIN SPEC 92001-1:2019 [11] defines verification for AI modules as "activity to ensure that an AI module is built correctly according to its specification". According to ISO/IEC DIS 24029-2:2022 [14], robustness specifications typically represent different conditions that can naturally or adversarially change in the domain in which the neural network is deployed. This includes natural variations as well as adversarial inputs encountered in the deployment environment. Therefore, neural network robustness verification is generally achieved by checking that the outputs are correct under abnormal inputs. Robustness verification is a subset of neural network verification that focuses specifically on verifying the robustness properties of the neural network. In neural network verification, the neural network F, input data set C, and property to be verified P are considered. The property P specifies the conditions that the output of the network must satisfy for inputs from any point in C [82]. Robustness verification aims to ensure robustness conditions such as the neural network produces correct output results or maintains a very low probability of incorrect outputs, regardless of the perturbations or attack methods applied, as long as the inputs remain within a specified bound. In this context, the input variation range within a specified bound is the input data set C, and property P is defined by the robustness condition. In robustness bounds verification, this condition may involve ensuring that the model's output remains unchanged or that the variation in output is below a certain threshold. In probabilistic robustness verification, the robustness condition is the probability of an erroneous output is below a specified threshold. Verification methods are widely studied for assessing the robustness of neural networks, particularly in terms of local robustness against adversarial attacks. Local robustness verification is more common than global robustness because they are easier to specify.

Formal verification, also known as reachability analysis or output reachable set estimation [83-86], was the earliest approaches used in robustness verification. In this approach, robustness verification is formulated as a satisfiability problem, where the goal is to identify the smallest set of inputs that satisfy the robustness condition and result a bound for local robustness. Other methods such as sensitivity analysis [87], stability analysis or stability region verification [88, 89], and safety region verification [27-30, 90-92] have also been proposed as robustness verification techniques. ISO/IEC DIS 24029-2:2022 [14] categorizes the robustness verification of neural networks into stability, sensitivity, relevance, and reachability. The formal representation of some properties in robustness verification, including adversarial samples, local robustness, output reachability, interval attributes, and Lipschitz attributes, have been comprehensively reviewed by Bu et al. [31], who also explore the relationships among these properties.

Statistical verification is another method to assess the local robustness. These approaches quantify the local robustness of neural networks by evaluating the probability of inputs satisfying or violating the robustness condition within the verification region. This methodology, sometimes referred to as probabilistic or quantitative verification/certification, can be implemented within a formal framework [69, 74, 85, 93, 94] or based on input domain sampling [29, 51, 95]. Statistical verification offers a way to address the challenges of formal verification while maintaining compatibility with its principles.

Researchers also refer to formal verification and statistical verification as qualitative and quantitative methods respectively [96]. Formal verification provides a binary judgment of whether robustness is achieved within the region, while statistical verification methods offer a quantitative measure of robustness by providing the probability of satisfying the robustness condition in the region.

*5.1.2 Formal Verification*

Formal verification can be complete or incomplete. Complete verification provides deterministic guarantees for all inputs within the bounds, while incomplete verification offers unilateral guarantees by the lower bounds.

*5.1.2.1 Complete Methods*

The earliest methods for neural network verification aimed to obtain exact solutions for robustness boundaries. These methods, referred to as complete verification by Li et al. [97], can determine precise bounds that would ensure the model's output remains unchanged within a specified range of perturbations. If the perturbations exceeded this range, the output could change, indicating an incorrect prediction. Techniques such as Satisfiability Modulo Theories (SMT) or Mixed Integer Linear Programming (MILP) were commonly used in these methods to rigorously and theoretically verify the neural network robustness to all possible inputs [67].

Katz et al. conducted pioneering studies for formal verification of neural networks. They demonstrated that verifying properties of DNNs with ReLU activation functions is an NP-complete problem in the paper [27], and developed Reluplex solver on the basis of SMT for the formal verification of neural networks. This method represents ReLU networks as piecewise linear functions and analyzes robustness by modeling the range of input variations as a hypersphere with a perturbation radius measured by $l_p$-norm (specifically, the $l_\infty$-norm in the case of Reluplex). This approach forms the basis for current methods used in robustness verification of neural networks. In a subsequent paper [98], they explored the scalability of Reluplex and highlighted that computational efficiency can be greatly enhanced by meticulously defining the properties to be verified and by parallelizing the processing. They proposed another framework Marabou [99] which supports any piecewise linear activation function as well as convolutional neural networks.

Several papers [67, 97] have reviewed complete formal verification methods for neural network robustness. However, these methods suffer from limitations. Each method is typically tailored to some specific network architectures and activation functions, such as feed-forward multilayer neural networks in paper [90], 2-class neural networks in paper [100], piecewise linear neural networks in paper [101], and ReLU activation functions in paper [60, 61, 83]. While MILP-based methods offer improved solution efficiency and scalability compared to SMT approaches, there are still two primary challenges in achieving complete verification of robustness bounds. Firstly, the computational complexity remains high, making exact verification feasible only for small-scale neural networks. Secondly, the MILP method is primarily applicable to piecewise linear neural networks, which restricts the network architecture to a limited set of nonlinear forms, typically involving pooling layers and piecewise linear activation functions.

*5.1.2.2 Incomplete Methods*

To address the limitations of complete verification, researchers have proposed incomplete verification methods [97]. These methods aim to compute a lower bound as an approximation to the exact bound. The lower bound ensures that inputs within the verified region will result in consistent outputs, thus guaranteeing robust behavior. In the process of calculating the output bound of the nonlinear activation function, over-approximation is used, resulting in a smaller radius than the exact minimum adversarial distortion. Therefore, it is important for approximation methods to derive a non-trivial (i.e., non-zero) lower bound and tighten it as much as possible towards the exact solution. This serves as an important criterion for evaluating the effectiveness of an incomplete verification method.

Ji et al. [67] categorized existing methods into several types based on the specific technique employed, including convex relaxation-based methods, abstract interpretation-based methods, Lipschitz constant-based methods, randomized smoothing-based methods, interval boundary propagation-based methods, and cybernetic theory-based methods.

Convex relaxation-based methods approximate the behavior of a neural network using a simpler convex function to simplify the proof process. These methods transform the problem of solving robustness bounds into a convex optimization problem [102-105]. Techniques like Semi-Definite Programming (SDP) [106, 107] are commonly employed in this approach. The linear and quadratic functions are used to constrain the activation functions of DNN, some typical methods are CROWN [63], DEEPG [80], and Fast-Lin[66]. Boopathy et al.[62] proposed the CNN-Cert tool for CNN robustness verification, can handle various CNN architectures, including convolutional layers, pooling layers, batch normalization layers, and residual blocks.

Abstract interpretation-based methods provide a more generalized framework for analyzing and proving robustness. These methods adopt an abstract approach to remove irrelevant details among all possible inputs, and use abstract domains to approximate the input, hidden, and output layers of a neural network. Abstract interpretation [108] techniques can be applied without considering the structure of a network as it acts on the data. Representative methods include AI2 [109], DeepZ [110], DiffAI [111], DeepPoly [112], feature-convex [113], and RefineZono [65].

Lipschitz constant-based methods estimate the Lipschitz constant to quantify how sensitive the model's outputs are to small changes in the input data, and thus estimates the lower bound of robustness behavior. A smaller Lipschitz constant indicates that the model is less affected by changes in the input, suggesting higher robustness. This method is applicable to various network architectures, bur one drawback is that Lipschitz constant-based methods tend to yield broader lower bounds that far from the exact solution. Notable methods in this category include Fast-Lip [66], CLEVER [114], DeepGO [84], LiPopt [64].

Randomized smoothing-based methods analyze robustness by introducing random noise to the input to observe its effect on the output. This method is initially introduced by Lecuyer et al. [115], they used Monte Carlo sampling to generate samples around the input data and proposed a robustness guarantee method for top-1 prediction. Subsequent studies by Li et al. [116], Cohen et al. [117], Xie et al. [118], Dhillon et al. [119] and Pinot et al. [120] further extended the randomized smoothing approach. These studies focused on different noise distributions, such as Gaussian, uniform, polynomial, and exponential, respectively.

Interval Bound Propagation (IBP)-based methods are used to calculate the output boundaries of a neural network for a given input range. These methods provide robustness guarantees related to different input constraints, such as the range of perturbations caused by adversarial attacks. This method was firstly proposed by Gowal et al. [121], they represented robustness bounds based on the $l_\infty$-norm and adopted the IBP method on the image classification model. Other notable methods derived from IBP include CROWN-IBP [122] and ReluVal [91].

Cybernetics-based methods [123-125] combine neural network robustness verification with robust control theory in a control loop framework. This method is generalized and does not require the neural network to satisfy differentiability or continuity.

*5.1.2.3 Strengths and Limitations*

Formal methods provide theoretical, rigorous, and complete robustness guarantees based on mathematical proofs of the correctness of neural network outputs. It ensures that all possible inputs within a given perturbation range satisfy the safety specification or so-called robustness condition, or that attacks within the verified regions are bound to be unsuccessful, which is useful in safety-critical applications where the effects of failure are severe. However, it is difficult to apply in practice. Leino et al. [36] pointed out that most techniques for verifying local robustness are expensive even on small models. The main shortcomings of formal verification evaluation methods can be summarized as follows:

a) Computationally inefficient, especially for large neural networks and high-dimensional input spaces, and therefore difficult to scale up to large models for practical applications;
b) Poor generalization, limited by activation function and network architectures that can be handled;
c) Provides qualitative robustness evaluations with no guarantees for perturbations outside the verification range.

*5.1.3 Statistical Verification*

Statistical verification gives the probabilistic robustness of the verified region by estimating the probability of inputs violating the robustness condition using sampling and statistical techniques. It extends the scope of verification to a probabilistic setting, considering the possibility of erroneous behavior with a certain probability. By allowing for probabilistic errors, it provides a larger bound with high confidence, such as a certified region with 99% or 99.99% probability of robustness [51, 74]. There are two main categories of statistical verification methods: white-box methods based on formal frameworks and black-box methods based on input layer sampling. White-box methods sample at the middle layer of the network, while black-box methods sample at the input layer. The method of input sampling should assume specific perturbation distributions, and is commonly used to assess robustness against data corruptions.

*5.1.3.1 Status of Development*

The concept of probabilistic verification was introduced by Mangal et al. [68], they argued that the traditional worst-case analysis of robustness against adversarial inputs is difficult to be satisfied by actual neural networks, and suggested a probabilistic robustness that exhibit at least $(1 - \epsilon)$ correctness to inputs within a $\delta$-neighborhood of original point. This paper proposed an approach to estimate the probability of robustness by abstract interpretation and importance sampling techniques.

Weng et al. [74] proposed the probabilistic robustness certification framework, PROVEN, to give a statistical guarantees for any bounded $l_p$ perturbations. It provides probabilistic bounds in closed form and allows for tighter robustness lower bounds. This is a typical white-box approach based on formal verification framework such as Fast-Lin [66], CROWN [63], and CNN-Cert [62].

Webb et al. [69] proposed the AMLS-based framework, which approximates the integral of perturbed inputs using the Monte Carlo method by counting the number of perturbed inputs. This method provides both formal verification, indicating the existence of counterexamples when a violation is detected, and statistical estimation when no violation is detected. If the probability of violation is below a given threshold, it is considered to be 0, and the verified perturbation range is approximated as the formal verifiable region.

Baluta et al. [93] propose NPAQ for quantitatively verifying neural network properties in a probably approximately correct (PAC) style, specifically designed for binarized neural networks (BNNs). And then they introduced PROVERO [94], applying PAC robustness to a wider network. This method is input layer sampling-based, so the verification process requires only black-box access to the DNNs.

Li et al. [126] performed robustness analysis on an approximate PAC model of the original DNN model, and proposed DeepPAC, which verifies the robustness of the PAC model and thus the robustness of the original model under a given confidence and error rate. Comparing with formal methods such as ERAN [112] and statistical methods like PROVERO [94], DeepPAC achieves higher robustness probabilities and larger verification radius.

Huang et al. [29] introduced the concept of ε-weakened robustness to quantify the probabilistic robustness, denotes an input region in which the proportion of counter inputs is less than a given ε-threshold value.

Levy and Katz [95] proposed a method for measuring and evaluating the robustness of neural networks to adversarial inputs, RoMA, which is also a black-box method that does not require knowledge of the design or weights of the network.

Anderson and Sojoudi [127] shift the focus from adversarial input to random uncertain input, the case where the input noise obeys any probability distribution. In another study [128], they introduce a data-driven probabilistic assessment method to offer high probability robustness certification for input with general distributions. Fazlyab et al. [85] also studied the statistical verification of neural networks when the input is perturbed by random perturbations with known mean and covariance. Huang et al. [51] investigated the robustness against random noise and other semantic perturbations. and proposed to obtain the maximum perturbation range through bisection search and sampling, applicable to various architectures and data perturbations.

Calafiore and Campi [129] proposed a framework for solving probabilistic robustness based on robust control analysis, and also provided an efficient number of samples required to achieve a specified probability of robustness.

Cardelli et al. [130] proposed a probabilistic robustness analysis method based on Bayesian inference and Gaussian processes. This approach aims to provide probabilistic guarantees for regions where no adversarial samples are detected. They also proposed a probabilistic robustness metric for Bayesian neural networks in the paper [131]. Wicker et al. [30] also studied the probabilistic safety of Bayesian neural networks against input perturbations. These studies provide effective methods for probabilistic robustness analysis of nondeterministic neural networks.

*5.1.3.2 Strengths and Limitations*

Statistical verification methods offer robustness verification across a defined input range by employing extensive sampling to ensure the neural network complies with the robustness condition with high probability. While providing only a degree of confidence in robustness rather than exact bounds, it is more practical, with following advantages:

a) Marked enhancement in evaluation efficiency, enabling scalability to large models;
b) Can provide a robustness measurement for perturbed regions outside the boundaries of formal verification, and can be used for robustness assessment of non-adversarial perturbations.

Nevertheless, existing statistical verification methods for probabilistic robustness still encounter some challenges:

a) Current methods predominantly focus on solving confidence bounds for local robustness, but sampling-based techniques may require large amounts of data to obtain valid decision bounds;

b) Assumptions about perturbation distributions in the input space are prevalent in existing methods. However, given the typically high input dimension of neural networks, accurately modeling perturbation scenarios becomes challenging, thus impacting the accuracy of the evaluation results;

c) The accuracy of probability estimates for robustness within the verification region is also affected by the adequacy and appropriateness of sampling strategies.

## 5.2 Robustness Testing Assessment

### 5.2.1 Adversarial Testing

Adversarial testing generates perturbed test cases based on adversarial attack techniques to evaluate the ability of neural network models to defend against adversarial attacks. By generating minimum adversarial examples, an estimation of the upper-bound of robustness can be obtained.

#### 5.2.1.1 Status of Development

Fawzi et al. [132] presented a framework for analyzing the robustness of classifiers against adversarial perturbations, proposing a general upper bound for adversarial robustness that is independent of the training process, highlighting that this metric and evaluation method are suited for assessing the robustness of already trained classifiers rather than the learning algorithms. The pointwise robustness defined by Bastani et al. [39] in section 3.2.3 is also represented by the lower bound of adversarial perturbations to estimate the smallest upper bound of local robustness.

Adversarial example is a typical perturbation for neural networks. Szegedy et al. [3] firstly reported the presence of adversarial examples in 2014, demonstrating the imperceptible perturbations to inputs could lead to incorrect behaviors of models. DIN SPEC 92001-2:2020 [4] defines the concepts of adversarial attacks and examples, adversarial attack is an algorithm developed by adversaries, returning adversarial perturbations or examples, and adversarial example is an input designed intentionally by an attacker to cause the model to make a mistake. Szegedy et al. [3] explained the existence of adversarial examples as being due to discontinuities in piecewise linear functions, inadequate training, or insufficient regularization resulting in overfitting. The research by Goodfellow et al. [133] suggested that linearity in high-dimensional spaces is a primary cause of adversarial examples, indicating that the vulnerability of deep neural network models to adversarial attacks mainly due to their linear part, and by transforming the model to be non-linear could reduce their vulnerability to adversarial attacks. In conclusion, the properties such as nonlinearity, discontinuity, and nonconvexity of neural networks [134] cause the solution of minimum adversarial samples not to be a convex optimization problem [39]. Therefore, the generation of adversarial examples and the approximation of the minimum value need to be solved in the assessment of robustness upper bounds based on adversarial testing. In the algorithm of adversarial attacks, the adversarial example corresponding to an input point $x$ can be represented as follows [39]:

$$\min[D(x, x+\delta)] \text{ subject to } F(x+\delta) = t, x+\delta \in [0,1]^n$$

where $F$ denotes the neural network model, $\delta$ is the perturbation added to the sample $x$, D is the distance metric to evaluate the perturbation degree, and $t$ is the targeted erroneous label. The constraint $F(x+\delta) = t$ is highly nonlinear, making direct solutions challenging. Current techniques simplify this solving problem by conducting thorough, layer-by-layer searches around the input point's neighborhood or encoding the network as a set of constraints [39]. Researchers have proposed numerous viable techniques and tools for generating adversarial examples, which can be classified into white-box and black-box techniques based on whether they require internal information about the network architecture and gradients [135, 136]. Typical white-box adversarial attack techniques include L-BFGS [3], FGSM [133], I-FGSM [137], BIM [138], PGD [46], JSMA [139], DeepFool [38], and C&W [78], while typical black-box adversarial attack techniques include ONE-PIXEL [140], MI-FGSM [141], UPSET [142], Adversarial Patch [143], and Houdini [144].

Goodfellow et al. [133] proposed the Fast Gradient Sign Method (FGSM) for generating adversarial examples which identifies the direction of the greatest gradient change within deep neural networks and adds a fixed-size perturbation in this direction to generate adversarial examples. FGSM is a quickly method but has a lower success rate. To improve the success rate, Kurakin et al. [138] proposed the Basic Iterative Method (BIM), generated adversarial examples are restricted to the effective region by the cropping function. Madry et al. [46] proposed the Projected Gradient Descent (PGD) adversarial attack algorithm, which incorporates random initialization and increases the number of iterations to enhance the effectiveness of the attack. Papernot et al. [139] developed the Jacobian-based Saliency Map Attack (JSMA) algorithm, it calculates the Jacobian matrix of the input sample to obtain a saliency map to indicate the most significant pixel positions to add perturbations.

Moosavi-Dezfooli et al. [38] proposed an effective method for approximating adversarial examples, DeepFool, the minimum distance of perturbation represents the local robustness upper bound. It measures the global robustness by calculating the average ratio of the minimum perturbation value to the input size.

Carlini and Wagner [78] developed an algorithm for generating adversarial examples that are closer to the original input, particularly for models with adversarial defenses. They choose the most suitable algorithm based on a combination of methods for determining the attack target class, constraint adjustment function, and box constraint methods. The selection is evaluated using standard deviation and probability metrics of the average distance of the generated adversarial examples. Since the generated adversarial samples are closer to the original samples, the robustness estimation results obtained will be more accurate.

Other methods include the ONE-PIXEL method by Su et al. [140] which changes only one pixel to generate adversarial examples, and the MI-FGSM proposed by Dong et al. [141], which is a black-box attack method with targeted objectives inspired by the I-FGSM and ILCM algorithms. Sarkar et al. [142] proposed UPSET base on adversarial generative networks for targeted black-box adversarial example generation. Brown et al. [143] proposed Adversarial Patch method, adding adversarial graphics on the original image to realize the attack. Cisse et al. [144] introduced the Houdini algorithm by dividing the loss function into random extremum and task loss parts.

Christensen et al. [145] presented the Adversarial Pivotal Tuning method, a framework for generating highly expressive adversarial manipulations or deviations from the training data. It allows for semantically manipulating images in a detailed, diverse, and photorealistic manner while preserving the original class. These manipulated images can be used for testing adversarial robustness.

Eleftheriadis et al. [146] proposed an adversarial robustness evaluation benchmark of attacks that includes representative cases of the most effective attacks for various aspects of robustness. They fine-tuned attack parameters to maximize success rates with minimum data perturbations. This benchmark can be used to assess robustness and the effectiveness of robust defenses.

*5.2.1.2 Strengths and Limitations*

Adversarial attack and defense techniques are mainly based on adversarial robustness analysis and propose effective defense methods to improve the robustness in the face of adversarial perturbations deliberately generated by attackers. Although the problem of robustness assessment can be solved by adversarial testing, the robustness upper bound obtained by this method can only provide a robustness certification that the inputs outside will fail, without effective guarantees for robustness within that bound. Moreover, the upper bound relies on the optimization of attack algorithms to estimate the minimum value of adversarial distance. However, due to over-approximation and simplification during the optimization process, the resulting upper bounds are too loose to provide a meaningful robustness assessment.

*5.2.2 Benchmark Testing*

Benchmark testing involves generating a benchmark dataset by applying perturbations to the pre-existing test set that measures correctness. This dataset encompasses various types and degrees of perturbations, allowing for a comprehensive evaluation of robustness. This method is suitable for neural network models with semantic features in their inputs, by specifying valid values for input features expected in real-world deployments, global robustness can be measured [14]. It can assess both adversarial robustness (AR) and corruption robustness (CR). The robustness is measured by performance indicators of the benchmark, with common metrics for classification models including confusion matrices, accuracy, F-measure, and AUC [147], more indicators detailed in section 4.1.2.2. Benchmark testing provides semantic perturbation-oriented robustness evaluation and is commonly used for assessing global robustness.

*5.2.2.1 Status of Development*

An intuitive way to measure robustness is to check the correctness of the model in the presence of perturbations, and a robust model should exhibit good performance even when faced with perturbations [101]. Therefore, researchers have proposed to measure robustness by the performance indicators on perturbed datasets, that is the benchmark testing assessment method. Xiao et al. [79] emphasize the importance of including a diverse test set that represents potential adversarial perturbations encountered in real-world scenarios when evaluating adversarial robustness.

Dong et al. [41] developed a benchmark for evaluating the adversarial robustness of image classifiers. They categorized adversarial attacks based on the knowledge of the attack into white-box attacks, output-based black-box attacks, and transferable black-box attacks, by the operation of the attack into targeted and untargeted attacks, and by perturbation metrics into $l_2$ and $l_\infty$ attacks. The test accuracy was used to measure robustness, and two complementary robustness curves to present the results. Croce et al. [148] pointed out that the attack and defense methods selected by Dong et al. [41] were not the best performing methods, and in some cases, the robustness was overestimated. They proposed a new benchmark for evaluating the adversarial robustness of image classifiers, using their adaptive method AutoAttack [149] to assess the impact of adversarial perturbations on model performance. Pintor et al. [150] designed a set of adversarial attack patches added to original images to evaluate the impact of adversarial attacks on real-world image recognition, constructing an adversarial robustness testing benchmark set. Carlini et al. [151] used a diverse set of adversarial attacks to measure model adversarial robustness and developed a robustness evaluation checklist. Ma et al. [152] introduced multi-granularity adversarial robustness testing metrics for deep learning, selecting 4 common adversarial attack algorithms to generate test sets, and measuring test sufficiency with neuron coverage and neural network layer coverage rates. Ling et al. [153] conducted research on the relationship between adversarial attacks and defenses in convolutional neural network models. They evaluated the robustness by combining 16 types of adversarial attacks and 10 attack capability indicators, as well as 13 types of adversarial defenses and 5 defense capability indicators to assess their effectiveness.

Hendrycks and Dietterich [40] proposed the ImageNet-C and ImageNet-P datasets as benchmarks for testing data corruption robustness for ImageNet classifiers. The benchmark includes 15 common types of image corruptions from noise, blur, weather, and digital categories, each with 5 severity levels. Kar et al.[154] aimed to obtain more realistic corrupt images by their 3d features, introducing 3d-specific corruptions such as depth of field to generate 3d image distortions.

*5.2.2.2 Strengths and Limitations*

Benchmark testing methods can provide a global assessment of neural network robustness, and are easy to implement, independent of the neural network architectures and activation functions, so can be applied to DNNs of any scale. Test samples with different perturbation types can be generated to evaluate the performance of the model under different conditions, such as covering various adversarial attack algorithms to evaluate adversarial robustness, and evaluating the model's ability to deal with real-world scenarios on datasets with different variations or corruptions. However, existing benchmark testing methods have many issues and shortcomings to be addressed, including:

a) Benchmark testing evaluation methods rely on well-defined robustness objectives and comprehensive test datasets, with robustness estimations varying due to differences in test datasets [13];

b) The composition of benchmark lacks theoretical guidance, especially the setting of the perturbation level is arbitrary, which makes the test set poorly representative and insufficient, and affects the credibility of the robustness assessment results;

c) The generation of perturbed test inputs is usually based on the parameters of the generation algorithm to control the degree of perturbation, which has little to do with the degree of variation in the image;

d) Generating various types and degrees of perturbed samples for each original sample leads to high test redundancy. Mature functional and performance testing benchmarks have not yet been developed, and the fairness and authority of testing standards need to be improved [155];

e) The correctness of the robustness benchmark is more relevant to the model's generalization capability, whereas robustness focuses less on the overall performance of the model on the entire test set and more on the correctness of the output for each test case. It is important to note that the performance indicators does not directly describe the robustness itself;

f) Generating robustness benchmarks based on the testing set also suffers from the problem of mischaracterizing typical operations, since robustness describes the performance of a model in the presence of deviations between the training and usage environments, and therefore the instances involved in the training process are samples of its typical operations.

*5.2.3 Test Adequacy Analysis*

*5.2.3.1 Adequacy Metrics Based on Neuron Coverage*

Pei et al. [156] firstly proposed neuron coverage (NC) as a driver for generating neural network test cases, providing a white-box testing method DeepXplore to optimize the neuron coverage rates. The neuron coverage is the ratio of the number of neurons activated by all test inputs to the total number of neurons in the DNN, and a neuron is considered activated if its output value exceeds a specified threshold.

Ma et al. [152] introduced the layer coverage for neural networks, considering the most active neurons and their combinations (or sequences) to develop metrics at different granularity, including:

1) Neuron-level coverage criteria:
   - k-multisection neuron coverage
   - neuron boundary coverage
   - strong neuron activation coverage
2) Layer-level coverage criteria:
   - top-k neuron coverage
   - top-k neuron patterns

Sun et al. [157] extended MC/DC coverage to DNNs, measuring neural network coverage based on changes in neurons' signs, values, or distances. Ma et al. also proposed coverage metrics for combinatorial testing of neural networks [158, 159], detecting the proportion of interactive neuron activations within a layer to check the combinational activation states of neurons in each layer.

ISO/IEC TR 29119-11: 2020 [10] for AI system testing guidelines summarizes existing white-box testing coverage metrics for neural networks, including:
- Neuron coverage
- Threshold coverage
- Sign change coverage
- Value change coverage
- Sign-sign coverage
- Layer coverage

Although numerous deep learning test coverage metrics can effectively guide the generation of test inputs, their efficacy in evaluating the robustness of neural networks has been questioned. The primary concern is related to adversarial robustness assessment, where a correlation between the detection of more adversarial examples and high neuron coverage rates is lacking [160, 161]. On the contrary, optimizing for coverage rate might even reduce the detection rate of adversarial examples [162]. Dong et al. [160] analyzed the relationship between neuron coverage rates and the robustness of neural network models, finding that higher neuron coverage does not necessarily translate into a better assessment of neural network model robustness. Li et al. [161] also suggested that the ability to detect incorrect predictions is due to the adversary-oriented search but not the high neural network structural coverage. Fabrice et al. [162] concluded from experiments that higher neuron coverage leads to fewer defects detected, fewer natural inputs, and more biased prediction preferences.

*5.2.3.2 Adequacy Metrics Based on Input Domain Coverage*

Beyond neuron coverage-related metrics, some researchers have proposed other types of adequacy analysis or coverage methods for deep learning models or AI systems.

Scenario coverage is an important metric for testing deep learning systems, especially when applied to complex AI systems. It plays a vital role in the testing of autonomous driving systems, generally based on the scenario description language [163-165].

Jiang et al. [166] proposed a data diversity metric for deep learning testing, using geometric diversity to measure the feature diversity of perturbed inputs.

Kim et al. [167] suggested measuring the quality and sufficiency of test generation by the degree of surprise in inputs, i.e., the behavioral difference between the test inputs and training data. They argued that good test inputs should be surprising but not overtly so compared to the training data.

# 6 Discussion

Neural network verification techniques for robustness assessment have made significant progress, providing various solutions and metrics for measuring robustness against adversarial attacks. As the impact of data corruptions on the behavior of neural networks has received more attention, the concept of corruption robustness has been widely mentioned. Data corruptions encompass a wide range of input anomalies that neural networks encounter in real-world scenarios. Testing methods have become a popular approach for assessing neural network robustness, especially for corruption robustness. They are preferred due to their simplicity, ease of implementation, and applicability to networks of any scale and structure, which is difficult with verification methods. Both methods have limitations, and in practice the testing method is applicable in most cases. In this section, we briefly summarize the challenges of both verification and testing methods and suggest some possible research directions.

## 6.1 Challenges

**Robustness Verification Assessment**. Formal verification and robustness boundary metrics are applicable to AR assessment, providing mathematical proofs for robustness in the regions within the boundary. However, these methods are difficult to apply even to medium scale neural networks, and the measurements based on maximum certified radius (or hypersphere bound) are not practical. Statistical verification of probabilistic robustness is more efficient and can be applied to both AR and CR. However, when it used for CR assessment, an important issue is that the perturbation range of $l_p$-ball is unsuitable for modeling data corruption. The hypersphere region is suitable for describing uncorrelated pixel changes in adversarial attacks on images, whereas image corruptions exhibit correlated changes among pixels, making hypersphere perturbation regions insufficient to represent local robustness. The bounds for such perturbations are more complex. In summary, existing boundary metrics based on hypersphere modeling are impractical, offering poor scalability in AR assessment and inefficiency in CR assessment.

**Robustness Testing Assessment**. The existing benchmark for assessing the robustness of neural networks is deficient in terms of representation. It exhibits limited coverage of perturbation types and levels, leading to a lack of confidence in the evaluation results regarding robustness. Additionally, the number of perturbed test inputs used for estimating robustness is constrained due to the associated costs and time limitations involved in testing. Consequently, a significant challenge lies in obtaining more accurate robustness evaluation results with a minimum number of test samples within the framework of testing-based assessment. Overall, the current benchmarks proposed for evaluating the robustness of neural networks still exhibit notable imperfections.

## 6.2 Future Directions

**Benchmark for robustness testing**. Developing standardized testing procedures and proposing effective benchmarks to test and assess robustness are crucial for testing assessment approach. Constructing a well-designed benchmark requires consideration of factors such as representativeness, repeatability, and comprehensiveness. A key research direction for future neural network robustness testing and assessment is to establish a benchmark library widely recognized by the robustness committee, providing a public, standardized evaluation platform. By collaboratively building and using benchmarks, we can better understand and improve the robustness of neural network models in the real world, thus promoting the application and development of technology.

**Integration of multiple assessment approaches**. A promising research direction is the exploration and integration of multiple assessment approaches. By combining various techniques, such as formal verification, adversarial testing, and benchmark methods, we can achieve a more comprehensive evaluation of neural network robustness. Formal verification techniques provide rigorous mathematical proofs and guarantees about the lower bounds of robustness. Adversarial testing and benchmark methods applying specially-designed perturbations to test the network's resilience. By synergistically leveraging the strengths of these different approaches, we can overcome their individual limitations and achieve a more holistic understanding of neural network robustness.

**Global robustness estimation based on local robustness**. Analyzing robustness from local to global is an effective robustness assessment method. Local robustness analysis verifies the output of perturbed samples within the neighborhood of a single local point. Combining the robustness of enough local points can estimate the global robustness. According to surveys, only a few papers have briefly discussed the feasibility of their methods for global robustness assessment in the experimental analysis section, and effective solutions for selecting sufficient local points to characterize global robustness and for adequately generalizing from local to global robustness are still lacking.

**Robustness evaluation of large-scale neural networks**. Several recent works have focused on the relationship between a neural network scale and its achieved robustness, and it has been consistently reported that larger networks help tremendously for robustness. Bubeck and Sellke [168] studied the law of robustness and proposed the tradeoff between the size of a model (as measured by the number of parameters) and its robustness (as measured by its Lipschitz constant). There is still a lack of effective methods for assessing the robustness of large models against attacks or various types of corruptions. Formal verification is difficult to scale to large models, and testing method is a promising solution.

# 7 Conclusion

This review aims to present a comprehensive overview of robustness assessment for neural networks applied to image recognition tasks. Two primary techniques, namely verification and testing, are extensively employed to support robustness assessment. Verification is predominantly utilized for evaluating adversarial robustness, while testing is commonly employed to assess corruption robustness. This paper provides an examination of the advancements made in neural network robustness assessment, encompassing robustness concepts, metrics, verification assessment methods, and testing assessment methods. The practical limitations inherent in each type of method are thoroughly discussed, leading to the assertion that testing assessment holds greater promise for future advancements. Additionally, the need for a comprehensive and standardized robustness evaluation benchmark is emphasized. This survey serves as a valuable resource for researchers and practitioners seeking to gain a deeper understanding of neural networks' robustness assessment in the context of image recognition tasks.

**REFERENCE**

[1] ISO/IEC, "Information technology — Artificial intelligence — Artificial intelligence concepts and terminology," 2021.

[2] ISO/IEC, "Software engineering - Systems and software Quality Requirements and Evaluation (SQuaRE) - Quality Model for AI-based systems," 2022.

[3] C. Szegedy, W. Zaremba, I. Sutskever, J. Bruna, D. Erhan, I. Goodfellow, and R. Fergus, "Intriguing properties of neural networks," ICLR, 2014.

[4] D. SPEC, "Artificial Intelligence — Life Cycle Processes and Quality Requirements — Part 2: Robustness," 2020.

[5] "Data corruption," https://en.wikipedia.org/wiki/Data_corruption.

[6] Y. Li, B. Xie, S. Guo, Y. Yang, and B. Xiao, "A Survey of Robustness and Safety of 2D and 3D Deep Learning Models against Adversarial Attacks," ACM Computing Surveys, vol. 56, no. 6, pp. 1-37, 2024-06-30, 2024.

[7] A. Serban, E. Poll, and J. Visser, "Adversarial Examples on Object Recognition: A Comprehensive Survey," ACM Computing Surveys, vol. 53, no. 3, pp. 1-38, 2021-05-31, 2021.

[8] ISO/IEC, "Information technology — Artificial intelligence — Overview of trustworthiness in artificial intelligence," 2020.

[9] ISO/IEC, "Framework for Artificial Intelligence (AI) Systems Using Machine Learning (ML)," 2021.

[10] ISO/IEC, "Software and systems engineering — Software testing — Part 11: Guidelines on the testing of AI-based systems," 2020.

[11] D. SPEC, "Artificial Intelligence — Life Cycle Processes and Quality Requirements —Part 1: Quality Meta Model," 2019.

[12] J. M. Zhang, M. Harman, L. Ma, and Y. Liu, "Machine Learning Testing: Survey, Landscapes and Horizons," IEEE Transactions on Software Engineering, pp. 1-1, 2020.

[13] ISO/IEC, "Artificial Intelligence (AI) — Assessment of the robustness of neural networks — Part 1: Overview," 2021.

[14] ISO/IEC, "Artificial intelligence (AI) — Assessment of the robustness of neural networks — Part 2: Methodology for the use of formal methods," 2022.

[15] M. Felderer, and R. Ramler, "Quality Assurance for AI-Based Systems: Overview and Challenges (Introduction to Interactive Session)," Software Quality: Future Perspectives on Software Engineering Quality, Swqd 2021, vol. 404, pp. 33-42, 2021.

[16] B. Jason. "Difference Between Algorithm and Model in Machine Learning.," https://machinelearningmastery.com/difference-between-algorithm-and-model-in-machine-learning/.

[17] ISO/IEC, "Systems and software engineering - Systems and software Quality Requirements and Evaluation (SQuaRE) - System and software quality models," 2011.

[18] IEEE, "IEEE Standard Glossary of Software Engineering Terminology," 2002.

[19] ISO/IEC, "Information technology — Artificial intelligence — Guidance on risk management," 2023.

[20] SAE, "Report on Unmanned Ground Vehicle Reliability," US-SAE, 2020.

[21] J. Fang, W. Liu, Y. Gao, Z. Liu, A. Zhang, X. Wang, and X. He, "Evaluating Post-hoc Explanations for Graph Neural Networks via Robustness Analysis," Advances in Neural Information Processing Systems, vol. 36, pp. 72446-72463, 2023-12-15, 2023.

[22] D. Hendrycks, and K. Gimpel, "A Baseline for Detecting Misclassified and Out-of-Distribution Examples in Neural Networks," 2017.

[23] S. Liang, Y. Li, and R. Srikant, "Enhancing The Reliability of Out-of-distribution Image Detection in Neural Networks."

[24] H. Barmer, R. Dzombak, M. Gaston, E. Heim, V. Palat, F. Redner, T. Smith, and N. Vanhoudnos, Robust and Secure AI, Software Engineering Institute, Carnegie Mellon University, 2022.

[25] IEC, "Guidance on software aspects of dependability," 2012.

[26] C.-H. Cheng, G. Nührenberg, C.-H. Huang, H. Ruess, and H. Yasuoka, "Towards dependability metrics for neural networks," in Proceedings of the 16th ACM-IEEE International Conference on Formal Methods and Models for System Design, Beijing, China, 2018, pp. 43–46.

[27] G. Katz, C. Barrett, D. L. Dill, K. Julian, and M. J. Kochenderfer, "Reluplex: An Efficient SMT Solver for VerifyingDeep Neural Networks," International Conference on Computer Aided Verification, pp. 97–117, 2017.

[28] W. Ruan, M. Wu, Y. Sun, X. Huang, and M. Kwiatkowska, "Global Robustness Evaluation of Deep Neural Networks with Provable Guarantees for the Hamming

Distance."

[29] P. Huang, Y. Yang, M. Liu, F. Jia, F. Ma, and J. Zhang, "ε-weakened robustness of deep neural networks," in Proceedings of the 31st ACM SIGSOFT International Symposium on Software Testing and Analysis, Virtual, South Korea, 2022, pp. 126–138.

[30] M. Wicker, L. Laurenti, A. Patane, and M. Kwiatkowska, "Probabilistic Safety for Bayesian Neural Networks," in Proceedings of the 36th Conference on Uncertainty in Artificial Intelligence (UAI), Proceedings of Machine Learning Research, 2020, pp. 1198--1207.

[31] L. Bu, L. Chen, Y. Dong, X. Huang, J. Li, and Q. Li, Research Progress and Trend of Formal Verification techniques for Artificial Intelligence systems, CCF Formalization Committee, 2019.

[32] S. Chen, H. Shen, R. Wang, and X. Wang, "Relationship Between Prediction Uncertainty and Adversarial Robustness," Ruan Jian Xue Bao/Journal of Software, vol. 33, no. 2, pp. 15, 2022.

[33] H. Zhang, Y. Yu, J. Jiao, E. Xing, L. El Ghaoui, and M. Jordan, "Theoretically principled trade-off between robustness and accuracy." pp. 7472-7482.

[34] J. Li, and G. Li, "The Triangular Trade-off between Robustness, Accuracy and Fairness in Deep Neural Networks: A Survey," ACM Computing Surveys, pp. 3645088, 2024-02-12, 2024.

[35] Y. Y. Yang, C. Rashtchian, H. Zhang, R. Salakhutdinov, and K. Chaudhuri, "A Closer Look at Accuracy vs. Robustness," 2020.

[36] K. Leino, Z. Wang, and M. Fredrikson, "Globally-robust neural networks." pp. 6212-6222.

[37] A. Kabaha, and D. Drachsler-Cohen, "Verification of Neural Networks' Global Robustness," 2024-03-06, 2024.

[38] S.-M. Moosavi-Dezfooli, A. Fawzi, and P. Frossard, "DeepFool: A Simple and Accurate Method to Fool Deep Neural Networks," in 2016 IEEE Conference on Computer Vision and Pattern Recognition (CVPR), 2016, pp. 2574-2582.

[39] O. Bastani, Y. Ioannou, L. Lampropoulos, D. Vytiniotis, A. V. Nori, and A. Criminisi, "Measuring Neural Net Robustness with Constraints," Advances in Neural Information Processing Systems 29 (Nips 2016), vol. 29, 2016.

[40] D. Hendrycks, and T. Dietterich, "Benchmarking Neural Network Robustness to Common Corruptions and Perturbations," 2019.

[41] Y. Dong, Q. A. Fu, X. Yang, T. Pang, H. Su, Z. Xiao, and J. Zhu, "Benchmarking Adversarial Robustness on Image Classification." pp. 318-328.

[42] Y. Zhang, Z. Wei, X. Zhang, and M. Sun, "Using Z3 for Formal Modeling and Verification of FNN Global Robustness," CoRR, vol. abs/2304.10558, /, 2023.

[43] F. Tramèr, A. Kurakin, N. Papernot, I. Goodfellow, D. Boneh, and P. Mcdaniel, "Ensemble Adversarial Training: Attacks and Defenses."

[44] N. Papernot, P. Mcdaniel, X. Wu, S. Jha, and A. Swami, "Distillation as a Defense to Adversarial Perturbations Against Deep Neural Networks."

[45] D. Meng, and H. Chen, "MagNet: A Two-Pronged Defense against Adversarial Examples," in Proceedings of the 2017 ACM SIGSAC Conference on Computer and Communications Security, Dallas, Texas, USA, 2017, pp. 135–147.

[46] A. Madry, A. Makelov, L. Schmidt, D. Tsipras, and A. Vladu, "Towards Deep Learning Models Resistant to Adversarial Attacks."

[47] F. Liao, M. Liang, Y. Dong, T. Pang, X. Hu, and J. Zhu, "Defense Against Adversarial Attacks Using High-Level Representation Guided Denoiser." pp. 1778-1787.

[48] Y. Li, M. Cheng, C.-J. Hsieh, and T. C. M. Lee, "A Review of Adversarial Attack and Defense for Classification Methods," The American Statistician, pp. 1-17, 2021.

[49] G. Hinton, O. Vinyals, and J. Dean, "Distilling the Knowledge in a Neural Network," Computer Science, vol. 14, no. 7, pp. 38-39, 2015.

[50] N. Das, M. Shanbhogue, S.-T. Chen, F. Hohman, S. Li, L. Chen, M. E. Kounavis, and D. H. Chau, "SHIELD: Fast, Practical Defense and Vaccination for Deep Learning using JPEG Compression," in Proceedings of the 24th ACM SIGKDD International Conference on Knowledge Discovery & Data Mining, London, United Kingdom, 2018, pp. 196–204.

[51] C. Huang, Z. Hu, X. Huang, and K. Pei, "Statistical Certification of Acceptable Robustness for Neural Networks," Artificial Neural Networks and Machine Learning – ICANN 2021. pp. 79-90.

[52] J. Mohapatra, T.-W. Weng, P.-Y. Chen, S. Liu, and L. Daniel, "Towards verifying robustness of neural networks against a family of semantic perturbations." pp. 244-252.

[53] A. Hamdi, and B. Ghanem, "Towards Analyzing Semantic Robustness of Deep Neural Networks," Computer Vision – ECCV 2020 Workshops. pp. 22-38.

[54] A. Hamdi, M. Müller, and B. Ghanem, "SADA: semantic adversarial diagnostic attacks for autonomous applications." pp. 10901-10908.

[55] M. A. Alcorn, Q. Li, Z. Gong, C. Wang, L. Mai, W.-S. Ku, and A. Nguyen, "Strike (with) a pose: Neural networks are easily fooled by strange poses of familiar objects." pp. 4845-4854.

[56] H.-T. D. Liu, M. Tao, C.-L. Li, D. Nowrouzezahrai, and A. Jacobson, "Beyond pixel norm-balls: Parametric adversaries using an analytically differentiable renderer."

[57] H. Qiu, C. Xiao, L. Yang, X. Yan, H. Lee, and B. Li, "SemanticAdv: Generating adversarial examples via attribute-conditioned image editing." pp. 19-37.

[58] X. Zeng, C. Liu, Y.-S. Wang, W. Qiu, L. Xie, Y.-W. Tai, C.-K. Tang, and A. L. Yuille, "Adversarial attacks beyond the image space." pp. 4302-4311.

[59] R. Ehlers, "Formal Verification of Piece-Wise Linear Feed-Forward Neural Networks." pp. 269-286.

[60] C.-H. Cheng, G. Nührenberg, and H. Ruess, "Maximum Resilience of Artificial Neural Networks," Automated Technology for Verification and Analysis. pp. 251-268.

[61] M. Fischetti, and J. Jo, "Deep neural networks as 0-1 mixed integer linear programs: A feasibility study," arXiv preprint arXiv:1712.06174, 2017.

[62] A. Boopathy, T.-W. Weng, P.-Y. Chen, S. Liu, and L. Daniel, "Cnn-cert: An efficient framework for certifying robustness of convolutional neural networks." pp. 3240-3247.

[63] H. Zhang, T.-W. Weng, P.-Y. Chen, C.-J. Hsieh, and L. Daniel, "Efficient neural network robustness certification with general activation functions," in Proceedings of the 32nd International Conference on Neural Information Processing Systems, Montréal, Canada, 2018, pp. 4944–4953.

[64] F. L. Gómez, P. Rolland, and V. Cevher, "Lipschitz constant estimation of Neural Networks via sparse polynomial optimization," 2020.

[65] G. Singh, T. Gehr, M. Püschel, and M. Vechev, "Boosting robustness certification of neural networks."

[66] L. Weng, H. Zhang, H. Chen, Z. Song, C.-J. Hsieh, L. Daniel, D. Boning, and I. Dhillon, "Towards Fast Computation of Certified Robustness for ReLU Networks," in Proceedings of the 35th International Conference on Machine Learning, Proceedings of Machine Learning Research, 2018, pp. 5276--5285.

[67] S. Ji, T. Du, S. Deng, P. Cheng, J. Shi, M. Yang, and B. Li, "Robustness Certification Research on Deep Learning Models: A Survey," CHINESE JOURNAL OF COMPUTERS, vol. 45, no. 1, pp. 17, 2022.

[68] R. Mangal, A. V. Nori, and A. Orso, "Robustness of Neural Networks: A Probabilistic and Practical Approach."

[69] S. Webb, T. Rainforth, Y. W. Teh, and M. P. Kumar, "A Statistical Approach to Assessing Neural Network Robustness."

[70] D. Gopinath, G. Katz, C. S. Pasareanu, and C. Barrett, "DeepSafe: A Data-driven Approach for Checking Adversarial Robustness in Neural Networks," 2017.

[71] G. Dong, J. Gao, R. Du, L. Tian, H. E. Stanley, and S. Havlin, "Robustness of network of networks under targeted attack," Physical Review E, vol. 87, no. 5, pp. 052804, 2013.

[72] G. Avila, T. Withers, and G. Holwell, "Retrospective risk assessment reveals likelihood of potential non-target attack and parasitism by Cotesia urabae (Hymenoptera: Braconidae): a comparison between laboratory and field-cage testing results," Biological Control, vol. 103, pp. 108-118, 2016.

[73] J. Jia, X. Cao, B. Wang, and N. Z. Gong, "Certified robustness for top-k predictions against adversarial perturbations via randomized smoothing," arXiv preprint arXiv:1912.09899, 2019.

[74] L. Weng, P.-Y. Chen, L. Nguyen, M. Squillante, A. Boopathy, I. Oseledets, and L. Daniel, "PROVEN: Verifying robustness of neural networks with a probabilistic approach." pp. 6727-6736.

[75] M. Wicker, X. Huang, and M. Kwiatkowska, "Feature-Guided Black-Box Safety Testing of Deep Neural Networks." pp. 408-426.

[76] D. G. Lowe, "Distinctive Image Features from Scale-Invariant Keypoints," International Journal of Computer Vision, vol. 60, no. 2, pp. 91-110, 2004/11/01, 2004.

[77] F. Croce, and M. Hein, "Provable robustness against all adversarial lp-perturbations for p≥1," CoRR, vol. abs/1905.11213, /, 2019.

[78] N. Carlini, and D. Wagner, "Towards Evaluating the Robustness of Neural Networks." pp. 39-57.

[79] C. Xiao, J.-Y. Zhu, B. Li, W. He, M. Liu, and D. Song, "Spatially transformed adversarial examples," arXiv preprint arXiv:1801.02612, 2018.

[80] M. Balunovic, M. Baader, G. Singh, T. Gehr, and M. Vechev, "Certifying geometric robustness of neural networks," Advances in Neural Information Processing Systems, vol. 32, 2019.

[81] ISO/IEC, "Information technology — Security techniques — Guidelines for the analysis and interpretation of digital evidence," 2015.

[82] "What is Neural Network Verification?," https://www.jiqizhixin.com/articles/2021-09-16-6.

[83] A. Lomuscio, and L. Maganti, "An approach to reachability analysis for feed-forward relu neural networks," arXiv preprint arXiv:1706.07351, 2017.

[84] W. Ruan, X. Huang, and M. Kwiatkowska, "Reachability analysis of deep neural networks with provable guarantees," arXiv preprint arXiv:1805.02242, 2018.

[85] M. Fazlyab, M. Morari, and G. J. Pappas, "Probabilistic verification and reachability analysis of neural networks via semidefinite programming." pp. 2726-2731.

[86] W. Xiang, H.-D. Tran, and T. T. Johnson, "Output reachable set estimation and verification for multilayer neural networks," IEEE transactions on neural networks and learning systems, vol. 29, no. 11, pp. 5777-5783, 2018.

[87] Y. Nie, Y. Wang, and M. Bansal, "Analyzing Compositionality-Sensitivity of NLI Models," 2018.

[88] H. Huang, Y. Qu, and H. X. Li, "Robust stability analysis of switched Hopfield neural networks with time-varying delay under uncertainty," Physics Letters A, vol. 345, no. 4-6, pp. 345-354, 2005.

[89] Y. Guo, "Globally robust stability analysis for stochastic Cohen–Grossberg neural networks with impulse control and time-varying delays," Ukrains' kyi Matematychnyi Zhurnal, vol. 69, no. 8, pp. 1049-1060, 2017.

[90] X. Huang, M. Kwiatkowska, S. Wang, and M. Wu, "Safety Verification of Deep Neural Networks," Computer Aided Verification. pp. 3-29.

[91] S. Wang, K. Pei, J. Whitehouse, J. Yang, and S. Jana, "Formal Security Analysis of Neural Networks using Symbolic Intervals," CoRR, vol. abs/1804.10829, /, 2018.

[92] S. Wang, K. Pei, J. Whitehouse, J. Yang, and S. Jana, "Efficient Formal Safety Analysis of Neural Networks," 2018, pp. 6369-6379.

[93] T. Baluta, S. Shen, S. Shinde, K. S. Meel, and P. Saxena, "Quantitative Verification of Neural Networks and Its Security Applications," in Proceedings of the 2019 ACM SIGSAC Conference on Computer and Communications Security, London, United Kingdom, 2019, pp. 1249–1264.

[94] T. Baluta, Z. L. Chua, K. S. Meel, and P. Saxena, "Scalable Quantitative Verification for Deep Neural Networks." pp. 312-323.

[95] N. Levy, and G. Katz, "RoMA: a Method for Neural Network Robustness Measurement and Assessment," arXiv preprint arXiv:2110.11088, 2021.

[96] Y. Zhang, Z. Zhao, G. Chen, F. Song, M. Zhang, T. Chen, and J. Sun, "QVIP: An ILP-based Formal Verification Approach for Quantized Neural Networks," in Proceedings of the 37th IEEE/ACM International Conference on Automated Software Engineering, Rochester, MI, USA, 2023, pp. Article 82.

[97] L. Li, T. Xie, and B. Li, "SoK: Certified Robustness for Deep Neural Networks." pp. 1289-1310.

[98] G. Katz, C. Barrett, D. L. Dill, K. Julian, and M. J. Kochenderfer, "Towards Proving the Adversarial Robustness of Deep Neural Networks," Electronic Proceedings in Theoretical Computer Science, vol. 257, pp. 19-26, 2017.

[99] G. Katz, D. A. Huang, D. Ibeling, K. Julian, C. Lazarus, R. Lim, P. Shah, S. Thakoor, H. Wu, A. Zeljić, D. L. Dill, M. J. Kochenderfer, and C. Barrett, "The Marabou Framework for Verification and Analysis of Deep Neural Networks," Computer Aided Verification. pp. 443-452.

[100] N. Narodytska, S. Kasiviswanathan, L. Ryzhyk, M. Sagiv, and T. Walsh, "Verifying Properties of Binarized Deep Neural Networks," Proceedings of the AAAI Conference on Artificial Intelligence, vol. 32, no. 1, 04/26, 2018.

[101] V. Tjeng, K. Xiao, and R. Tedrake, "Evaluating Robustness of Neural Networks with Mixed Integer Programming," ICLR, 2019.

[102] E. Wong, and Z. Kolter, "Provable Defenses against Adversarial Examples via the Convex Outer Adversarial Polytope," in Proceedings of the 35th International Conference on Machine Learning, Proceedings of Machine Learning Research, 2018, pp. 5286--5295.

[103] E. Wong, F. Schmidt, J. H. Metzen, and J. Z. Kolter, "Scaling provable adversarial defenses," Advances in Neural Information Processing Systems, vol. 31, 2018.

[104] K. Dvijotham, R. Stanforth, S. Gowal, T. A. Mann, and P. Kohli, "A Dual Approach to Scalable Verification of Deep Networks." p. 3.

[105] K. Dvijotham, M. Garnelo, A. Fawzi, and P. Kohli, "Verification of deep probabilistic models," arXiv preprint arXiv:1812.02795, 2018.

[106] A. Raghunathan, J. Steinhardt, and P. Liang, "Certified Defenses against Adversarial Examples," arXiv, 2018.

[107] A. Raghunathan, J. Steinhardt, and P. S. Liang, "Semidefinite relaxations for certifying robustness to adversarial examples," Advances in neural information processing systems, vol. 31, 2018.

[108] P. Cousot, and R. Cousot, "Abstract interpretation: a unified lattice model for static analysis of programs by construction or approximation of fixpoints," in Proceedings of the 4th ACM SIGACT-SIGPLAN symposium on Principles of programming languages, Los Angeles, California, 1977, pp. 238–252.

[109] T. Gehr, M. Mirman, D. Drachsler-Cohen, P. Tsankov, S. Chaudhuri, and M. Vechev, "Ai2: Safety and robustness certification of neural networks with abstract interpretation." pp. 3-18.

[110] G. Singh, T. Gehr, M. Mirman, M. Püschel, and M. Vechev, "Fast and effective robustness certification," in Proceedings of the 32nd International Conference on
31

Neural Information Processing Systems, Montréal, Canada, 2018, pp. 10825–10836.

[111] M. Mirman, T. Gehr, and M. Vechev, "Differentiable Abstract Interpretation for Provably Robust Neural Networks," in Proceedings of the 35th International Conference on Machine Learning, Proceedings of Machine Learning Research, 2018, pp. 3578--3586.

[112] G. Singh, T. Gehr, M. Püschel, and M. Vechev, "An abstract domain for certifying neural networks," Proc. ACM Program. Lang., vol. 3, no. POPL, pp. Article 41, 2019.

[113] S. Pfrommer, B. Anderson, J. Piet, and S. Sojoudi, "Asymmetric Certified Robustness via Feature-Convex Neural Networks," Advances in Neural Information Processing Systems, vol. 36, pp. 52365-52400, 2023-12-15, 2023.

[114] T. W. Weng, H. Zhang, P. Y. Chen, J. Yi, D. Su, Y. Gao, C. J. Hsieh, and L. Daniel, "Evaluating the Robustness of Neural Networks: An Extreme Value Theory Approach."

[115] M. Lecuyer, V. Atlidakis, R. Geambasu, D. Hsu, and S. Jana, "Certified robustness to adversarial examples with differential privacy." pp. 656-672.

[116] B. Li, C. Chen, W. Wang, and L. Carin, "Certified adversarial robustness with additive noise," Advances in neural information processing systems, vol. 32, 2019.

[117] J. Cohen, E. Rosenfeld, and Z. Kolter, "Certified Adversarial Robustness via Randomized Smoothing," in Proceedings of the 36th International Conference on Machine Learning, Proceedings of Machine Learning Research, 2019, pp. 1310--1320.

[118] C. Xie, J. Wang, Z. Zhang, Z. Ren, and A. Yuille, "Mitigating adversarial effects through randomization," arXiv preprint arXiv:1711.01991, 2017.

[119] G. S. Dhillon, K. Azizzadenesheli, Z. C. Lipton, J. Bernstein, J. Kossaifi, A. Khanna, and A. Anandkumar, "Stochastic activation pruning for robust adversarial defense," arXiv preprint arXiv:1803.01442, 2018.

[120] R. Pinot, L. Meunier, A. Araujo, H. Kashima, F. Yger, C. Gouy-Pailler, and J. Atif, "Theoretical evidence for adversarial robustness through randomization," Advances in neural information processing systems, vol. 32, 2019.

[121] S. Gowal, K. D. Dvijotham, R. Stanforth, R. Bunel, C. Qin, J. Uesato, R. Arandjelovic, T. Mann, and P. Kohli, "Scalable verified training for provably robust image classification." pp. 4842-4851.

[122] H. Zhang, H. Chen, C. Xiao, S. Gowal, R. Stanforth, B. Li, D. Boning, and C.-J. Hsieh, "Towards stable and efficient training of verifiably robust neural networks," arXiv preprint arXiv:1906.06316, 2019.

[123] Y.-S. Wang, T.-W. Weng, and L. Daniel, "Verification of neural network control policy under persistent adversarial perturbation," arXiv preprint arXiv:1908.06353, 2019.

[124] S. Carr, N. Jansen, and U. Topcu, "Verifiable RNN-based policies for POMDPs under temporal logic constraints," arXiv preprint arXiv:2002.05615, 2020.

[125] B. Wang, Z. Shi, and S. Osher, "Resnets ensemble via the feynman-kac formalism to improve natural and robust accuracies," Advances in Neural Information Processing Systems, vol. 32, 2019.

[126] R. Li, P. Yang, C. C. Huang, Y. Sun, B. Xue, and L. Zhang, "Towards Practical Robustness Analysis for DNNs based on PAC-Model Learning." pp. 2189-2201.

[127] B. G. Anderson, and S. Sojoudi, "Certifying neural network robustness to random input noise from samples," arXiv preprint arXiv:2010.07532, 2020.

[128] B. G. Anderson, and S. Sojoudi, "Data-Driven Assessment of Deep Neural Networks with Random Input Uncertainty," 2020.

[129] G. C. Calafiore, and M. C. Campi, "The scenario approach to robust control design," IEEE Transactions on Automatic Control, vol. 51, no. 5, pp. 742-753, 2006.

[130] L. Cardelli, M. Kwiatkowska, L. Laurenti, and A. Patane, "Robustness guarantees for Bayesian inference with Gaussian processes," in Proceedings of the Thirty-Third AAAI Conference on Artificial Intelligence and Thirty-First Innovative Applications of Artificial Intelligence Conference and Ninth AAAI Symposium on Educational Advances in Artificial Intelligence, Honolulu, Hawaii, USA, 2019, pp. Article 952.

[131] L. Cardelli, M. Kwiatkowska, L. Laurenti, N. Paoletti, A. Patane, and M. Wicker, "Statistical Guarantees for the Robustness of Bayesian Neural Networks." pp. 5693-5700.

[132] A. Fawzi, O. Fawzi, and P. Frossard, "Analysis of classifiers' robustness to adversarial perturbations," Machine Learning, vol. 107, no. 3, pp. 481-508, 2017.

[133] I. J. Goodfellow, J. Shlens, and C. Szegedy, "Explaining and Harnessing Adversarial Examples," in International Conference on Learning Representations (ICLR), 2015.

[134] G. F. Montufar, R. Pascanu, K. Cho, and Y. Bengio, "On the number of linear regions of deep neural networks," Advances in neural information processing systems,

vol. 27, 2014.

[135] W. Pan, X. Wang, M. Song, and C. Chen, "Survey on generating adversarial examples," Ruan Jian Xue Bao/Journal of Software, vol. 31, no. 1, pp. 15, 2020.

[136] H. Sun, J. Chen, L. Lei, K. Ji, and G. Kuang, "Adversarial robustness of deep convolutional neural network-based image recognition models: A review," Journal of Radars, vol. 10, no. 4, pp. 24, 2021.

[137] A. Kurakin, I. J. Goodfellow, and S. Bengio, "Adversarial examples in the physical world," Artificial intelligence safety and security, pp. 99-112: Chapman and Hall/CRC, 2018.

[138] A. Kurakin, I. Goodfellow, S. Bengio, Y. Dong, F. Liao, M. Liang, T. Pang, J. Zhu, X. Hu, and C. Xie, "Adversarial attacks and defences competition." pp. 195-231.

[139] N. Papernot, P. McDaniel, S. Jha, M. Fredrikson, Z. B. Celik, and A. Swami, "The limitations of deep learning in adversarial settings." pp. 372-387.

[140] J. Su, D. V. Vargas, and K. Sakurai, "One pixel attack for fooling deep neural networks," IEEE Transactions on Evolutionary Computation, vol. 23, no. 5, pp. 828-841, 2019.

[141] Y. Dong, F. Liao, T. Pang, H. Su, J. Zhu, X. Hu, and J. Li, "Boosting adversarial attacks with momentum." pp. 9185-9193.

[142] S. Sarkar, A. Bansal, U. Mahbub, and R. Chellappa, "UPSET and ANGRI: Breaking high performance image classifiers," arXiv preprint arXiv:1707.01159, 2017.

[143] T. B. Brown, D. Mané, A. Roy, M. Abadi, and J. Gilmer, "Adversarial patch," arXiv preprint arXiv:1712.09665, 2017.

[144] M. Cisse, Y. Adi, N. Neverova, and J. Keshet, "Houdini: Fooling deep structured prediction models," arXiv preprint arXiv:1707.05373, 2017.

[145] P. E. Christensen, V. Snæbjarnarson, A. Dittadi, S. Belongie, and S. Benaim, "Assessing Neural Network Robustness via Adversarial Pivotal Tuning." pp. 2952-2961.

[146] C. Eleftheriadis, A. Symeonidis, and P. Katsaros, "Adversarial robustness improvement for deep neural networks," Machine Vision and Applications, vol. 35, no. 3, pp. 35, 2024-03-14, 2024.

[147] N. Japkowicz, "Why question machine learning evaluation methods." pp. 6-11.

[148] F. Croce, M. Andriushchenko, V. Sehwag, E. Debenedetti, N. Flammarion, M. Chiang, P. Mittal, and M. Hein, "Robustbench: a standardized adversarial robustness benchmark," arXiv preprint arXiv:2010.09670, 2020].

[149] F. Croce, and M. Hein, "Reliable evaluation of adversarial robustness with an ensemble of diverse parameter-free attacks." pp. 2206-2216.

[150] M. Pintor, D. Angioni, A. Sotgiu, L. Demetrio, A. Demontis, B. Biggio, and F. Roli, "ImageNet-Patch: A dataset for benchmarking machine learning robustness against adversarial patches," Pattern Recognition, vol. 134, pp. 109064, 2023.

[151] N. Carlini, A. Athalye, N. Papernot, W. Brendel, J. Rauber, D. Tsipras, I. Goodfellow, A. Madry, and A. Kurakin, "On evaluating adversarial robustness," arXiv preprint arXiv:1902.06705, 2019.

[152] L. Ma, F. Juefei-Xu, F. Zhang, J. Sun, M. Xue, B. Li, C. Chen, T. Su, L. Li, Y. Liu, J. Zhao, and Y. Wang, "DeepGauge: multi-granularity testing criteria for deep learning systems," in Proceedings of the 33rd ACM/IEEE International Conference on Automated Software Engineering, 2018, pp. 120-131.

[153] X. Ling, S. Ji, J. Zou, J. Wang, C. Wu, B. Li, and T. Wang, "Deepsec: A uniform platform for security analysis of deep learning model." pp. 673-690.

[154] O. F. Kar, T. Yeo, A. Atanov, and A. Zamir, "3d common corruptions and data augmentation." pp. 18963-18974.

[155] White paper on artificial intelligence standardization, China Institute of Electronic Technology Standardization, 2021.

[156] K. Pei, Y. Cao, J. Yang, and S. Jana, "DeepXplore: Automated Whitebox Testing of Deep Learning Systems," in Proceedings of the 26th Symposium on Operating Systems Principles - SOSP '17, 2017, pp. 1-18.

[157] Y. Sun, X. Huang, D. Kroening, J. Sharp, M. Hill, and R. Ashmore, "Testing deep neural networks," arXiv preprint arXiv:1803.04792, 2018.

[158] L. Ma, F. Juefei-Xu, M. Xue, B. Li, and J. Zhao, "DeepCT: Tomographic Combinatorial Testing for Deep Learning Systems."

[159] L. Ma, F. Zhang, M. Xue, B. Li, Y. Liu, J. Zhao, and Y. Wang, "Combinatorial Testing for Deep Learning Systems," 2018.

[160] Y. Dong, P. Zhang, J. Wang, S. Liu, J. Sun, J. Hao, X. Wang, L. Wang, J. S. Dong, and D. Ting, "There is Limited Correlation between Coverage and Robustness for Deep Neural Networks," 2019].

[161] Z. Li, X. Ma, C. Xu, and C. Cao, "Structural Coverage Criteria for Neural Networks Could Be Misleading." pp. 89–92.

[162] F. Harel-Canada, L. Wang, M. A. Gulzar, Q. Gu, and M. Kim, "Is neuron coverage a meaningful measure for testing deep neural networks?," in Proceedings of the

28th ACM Joint Meeting on European Software Engineering Conference and Symposium on the Foundations of Software Engineering, Virtual Event, USA, 2020, pp. 851-862.

[163] A. Stranjak, I. Čavrak, and M. Žagar, "Scenario Description Language for Multi-agent Systems," New Frontiers in Applied Artificial Intelligence. pp. 855-864.

[164] X. Zhang, S. Khastgir, and P. Jennings, "Scenario Description Language for Automated Driving Systems: A Two Level Abstraction Approach." pp. 973-980.

[165] C.-H. Cheng, C.-H. Huang, and H. Yasuoka, "Quantitative Projection Coverage for Testing ML-enabled Autonomous Systems," Automated Technology for Verification and Analysis. pp. 126-142.

[166] Z. Jiang, H. Li, and R. Wang, "Efficient generation of valid test inputs for deep neural networks via gradient search," Journal of Software: Evolution and Process, pp. e2550, 2023.

[167] J. Kim, R. Feldt, and S. Yoo, "Guiding Deep Learning System Testing Using Surprise Adequacy," in 2019 IEEE/ACM 41st International Conference on Software Engineering (ICSE), 2019, pp. 1039-1049.

[168] S. Bubeck, and M. Sellke, "A Universal Law of Robustness via Isoperimetry," J. ACM, vol. 70, no. 2, pp. Article 10, 2023.